\documentclass[runningheads]{llncs}
\usepackage[T1]{fontenc}
\usepackage{graphicx}
\usepackage{subcaption}
\usepackage{amsmath}
\usepackage{hyperref}
\usepackage{multirow} 
\usepackage{booktabs}  

\usepackage{color}

\begin{document}
\title{QUEST: Quality-aware Semi-supervised Table Extraction for Business Documents}
\titlerunning{QUEST}
\author{Eliott THOMAS\inst{1,2}\orcidID{0009-0008-5266-8797} \and  
Mickael COUSTATY\inst{2}\orcidID{0000-0002-0123-439X} \and  
Aurélie JOSEPH\inst{1}\orcidID{0000-0002-5499-6355} \and  
Gaspar DELOIN\inst{1}\orcidID{0009-0007-2449-5385} \and  
Elodie CAREL\inst{1}\orcidID{0000-0002-2230-0018} \and  
Vincent POULAIN D'ANDECY\inst{1}\orcidID{0009-0008-5515-3561} \and  
Jean-Marc OGIER\inst{2}\orcidID{0000-0002-5666-475X}}  

\authorrunning{E. Thomas et al.}  

\institute{Yooz, France\\
\email{\{eliott.thomas,aurelie.joseph,gaspar.deloin,\\
elodie.carel,Vincent.PoulaindAndecy\}@getyooz.com} \and  
La Rochelle Université, France\\
\email{\{eliott.thomas,mickael.coustaty,jean-marc.ogier\}@univ-lr.fr}}   

\maketitle              % typeset the header of the contribution
\begin{abstract}

Automating table extraction (TE) from business documents is critical for industrial workflows but remains challenging due to sparse annotations and error-prone multi-stage pipelines. While semi-supervised learning (SSL) can leverage unlabeled data, existing methods rely on confidence scores that poorly reflect extraction quality. We propose QUEST, a Quality-aware Semi-supervised Table extraction framework designed for Business documents. QUEST introduces a novel quality assessment model that evaluates structural and contextual features of extracted tables, trained to predict F1 scores instead of relying on confidence metrics. This quality-aware approach guides pseudo-label selection during iterative SSL training, while diversity measures (DPP, Vendi score, IntDiv) mitigate confirmation bias. Experiments on a proprietary business dataset (1k annotated + 10k unannotated documents) show QUEST improves F1 from 64\% to 74\% and reduces empty predictions by 45\% (12\% to 6.5\%). On the DocILE benchmark (600 annotated + 20k unannotated documents), QUEST achieves a 50\% F1 score (up from 42\%) and reduces empty predictions by 19\% (27\% to 22\%). The framework’s interpretable quality assessments and robustness to annotation scarcity make it particularly suited for business documents, where structural consistency and data completeness are paramount.

\end{abstract}

\section{Introduction}  

Document intelligence systems are transforming modern business operations by automating the extraction of critical information from unstructured documents~\cite{Xu2020LayoutLMv2MP,Xu2019LayoutLMPO}. Within this domain, Table Extraction (TE) plays a pivotal role in converting tabular data into structured formats for downstream applications. Efficient TE is critical for processing business documents such as invoices, financial reports, and product catalogs, where structured tabular data drives operational workflows. Traditional approaches~\cite{Gao2019ICDAR2C,Gbel2012AMF,Gbel2013ICDAR2T,Smock2021PubTables1MTC} combining Table Detection (TD) and Table Structure Recognition (TSR) struggle with key challenges including format diversity across vendors~\cite{Gao2019ICDAR2C}, error propagation from element interdependence, and annotation scarcity due to document confidentiality~\cite{vSimsa2023DocILEBF}. These issues compound in industrial settings, where extraction failures disrupt downstream workflows like financial auditing. This is a problem explicitly highlighted in the DocILE benchmark’s analysis of real-world document processing pipelines~\cite{vSimsa2023DocILEBF}.

\begin{figure}[h]  
\centering  
\begin{subfigure}[b]{0.48\textwidth}  
    \centering  
    \includegraphics[width=\textwidth]{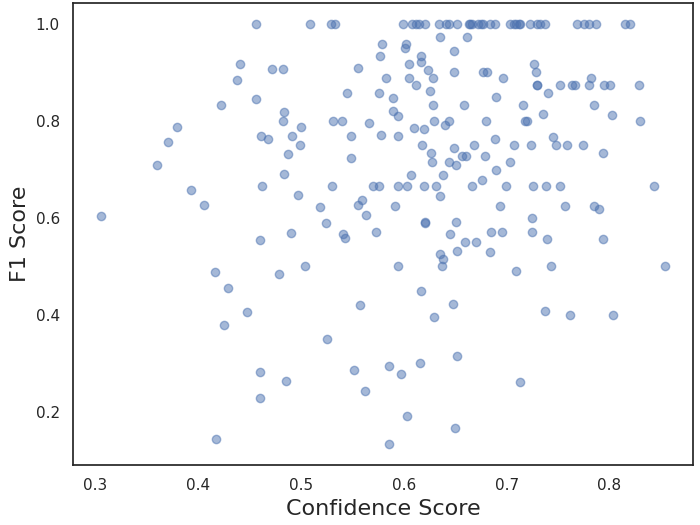}  
    \label{fig:confidence_vs_f1}  
\end{subfigure}  
\hfill  
\begin{subfigure}[b]{0.48\textwidth}  
    \centering  
    \includegraphics[width=\textwidth]{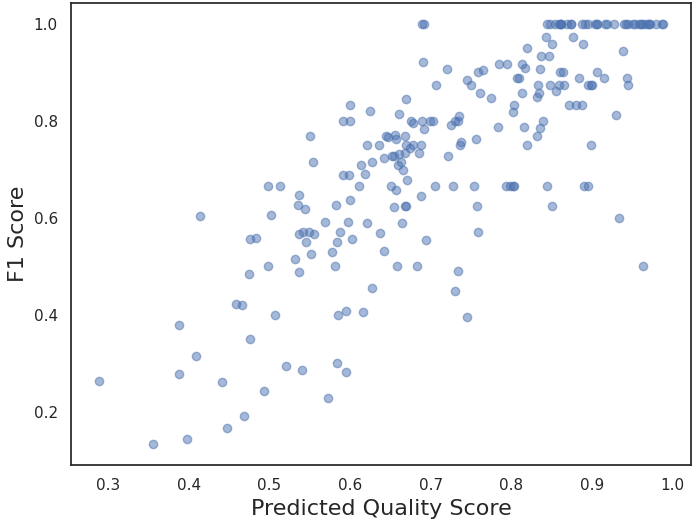}  
    \label{fig:quality_vs_f1}  
\end{subfigure}  
\caption{Correlation analysis (Pearson's $r$) with ground truth quality: Conventional confidence scores ($r=0.22$, RMSE=0.24) vs. QUEST ($r=0.80$, RMSE=0.13), demonstrating 3.6$\times$ stronger linear relationship and 46\% lower prediction error.}  
\label{fig:teaser}  
\end{figure}

To address annotation scarcity, semi-supervised learning (SSL) leverages unlabeled data through pseudo-label selection, reducing human annotation efforts. However, SSL effectiveness heavily depends on selecting high-quality pseudo-labels for training. Conventional SSL methods~\cite{Liu2021UnbiasedTF,Sohn2020FixMatchSS,Sohn2020ASS,Xu2021EndtoEndSO} rely on confidence scores that poorly correlate with actual extraction quality, as shown in Figure~\ref{fig:teaser}. This leads to two critical issues: incorporating erroneous predictions with high confidence into training, while discarding correct extractions with low confidence scores. Such selection errors are particularly damaging in business document analysis, where a single misplaced row or column can invalidate an entire table's structure.

We propose \textbf{QUEST: Quality-aware Semi-supervised Table Extraction for Business Documents}, a novel framework that fundamentally reimagines pseudo-label selection through verifiable quality assessment. Drawing inspiration from complexity theory~\cite{Arora2009ComputationalCA,Cook1971TheCO}, QUEST leverages a key insight: verifying the quality of an extracted table is more reliable than generating confidence scores during extraction. Our framework trains a specialized model to evaluate structural, layout and contextual features of extracted tables, producing interpretable F1 score predictions that directly measure extraction quality. This approach, combined with diversity measures (DPP~\cite{Kulesza2012DeterminantalPP}, Vendi score~\cite{Friedman2022TheVS}, IntDiv~\cite{Benhenda2017ChemGANCF}), ensures robust pseudo-label selection while preventing confirmation bias during iterative SSL training.  

Our key contributions are as follows:  
\begin{itemize}  
    \item \textbf{The QUEST Framework}: A quality-aware SSL method that revolutionizes business document extraction through interpretable quality assessment, evaluating both structural consistency and contextual plausibility of extracted tables. This approach bridges the gap between model confidence and actual extraction quality.  
    
    \item \textbf{Diversity-Guided Training}: A novel integration of complementary diversity metrics (DPP, Vendi scores, IntDiv) with quality-based selection, creating a robust pseudo-labeling strategy that systematically reduces error propagation in SSL iterations.  
    
    \item \textbf{Empirical Validation}: Comprehensive evaluation showing significant improvements: on our proprietary business dataset (1,000 annotated + 10,000 unannotated documents), QUEST boosts F1 from 64\% to 74\% while reducing empty predictions by 45\% (12\% to 6.5\%). On DocILE~\cite{vSimsa2023DocILEBF} (600 annotated + 20,000 unannotated), it achieves 50\% F1 (from 42\%) with 19\% fewer empty predictions (27\% to 22\%).  
\end{itemize}   

QUEST's modular design addresses the unique challenges of business documents, where structural consistency and data completeness are paramount. By emphasizing interpretable, feature-based assessments over raw confidence metrics, it reduces annotation overhead and avoids issues caused by unbalanced data. This approach yields transparent quality scores that enterprises can trust for operational workflows, while remaining adaptable to diverse tables.  

\section{Related Work}

\subsection{Table Extraction Methods}  

Business documents present unique challenges for table extraction systems, as evidenced in datasets such as DocILE~\cite{vSimsa2023DocILEBF}. These documents often contain diverse data types, inconsistent formatting, and arbitrary content placement, making accurate extraction particularly challenging~\cite{Shahab2010AnOA}. The scarcity of annotated data in business settings remains a significant constraint for supervised learning approaches~\cite{Schreiber2017DeepDeSRTDL}.  Table extraction has evolved from early heuristic approaches~\cite{Zanibbi2004ASO} to modern learning-based methods. This task typically involves two main components: Table Detection (TD)\cite{Gilani2017TableDU} and Table Structure Recognition (TSR)\cite{Chi2019ComplicatedTS}. Early approaches relied heavily on rule-based systems and geometric analysis, while recent deep learning architectures~\cite{Paliwal2019TableNetDL,Prasad2020CascadeTabNetAA} and transformer-based models have shown notable improvements by leveraging global context and reducing post-processing steps. However, these approaches typically require substantial labeled data, which is often unavailable in business settings.  
Recent advancements such as Table Transformer (TATR)\cite{Smock2021PubTables1MTC} have demonstrated strong performance by leveraging DETR-based architectures\cite{Carion2020EndtoEndOD} trained on table-specific datasets like PubTables-1M~\cite{Smock2021PubTables1MTC} and FinTabNet~\cite{Zheng2020GlobalTE}. Thomas et al.\cite{thomas2025raptorrefinedapproachproduct} address common extraction errors in business documents through specialized post-processing, highlighting the ongoing challenges when working with business tables. More efficient architectures like YOLOv9\cite{Wang2024YOLOv9LW}, chosen for its validated performance in production document analysis, offer improved speed-accuracy trade-offs, making them suitable for practical document analysis systems, but still require substantial labeled data for optimal performance. Evaluation metrics for these systems have also evolved, from simple structural measures like tree edit distance similarity (TEDS)~\cite{Zhong2019ImagebasedTR} and directed adjacency relations (DAR)~\cite{Gbel2012AMF,Gbel2013ICDAR2T}, to more comprehensive metrics like GRITS~\cite{Smock2022GriTSGT} that evaluates tables directly in their natural matrix form. GRITS offers three variants: GRITS-Top for topology recognition, GRITS-Con for content recognition, and GRITS-Loc for location recognition.

\subsection{Semi-Supervised Learning in Document Analysis}  
Semi-supervised learning (SSL)~\cite{Lee2013PseudoLabelT,Tarvainen2017MeanTA} has evolved from classification, assigning a single image label, to detection-specific frameworks~\cite{Sohn2020ASS}, requiring location and identification. Object detection's complexity, particularly for interconnected structures like tables, necessitates adapting SSL for spatial consistency of pseudo-labels and reliable object-level confidence measures. Although early SSL used multistage training~\cite{Liu2021UnbiasedTF}, recent end-to-end methods~\cite{Xu2021EndtoEndSO} show promise.  
In document analysis, model confidence has traditionally driven pseudo-label selection~\cite{Sohn2020FixMatchSS}, but its limitations are shown in several works~\cite{Huang2025InfantAction,Tang2021HumbleTT,Zhou2021InstantTeachingAE}. Recent table detection work~\cite{Ehsan2024EndtoEndSA} improves pseudo-label quality via novel matching strategies, but still primarily relies on confidence measures that poorly correlate with actual extraction quality. Building on insights from transfer learning approaches~\cite{Ruder2017LearningTS} that demonstrate the complementary value of diversity and similarity metrics, our work adapts this paradigm to the pseudo-label selection task.  
Addressing quality assessment, diversity measures like Determinantal Point Process~\cite{Kulesza2012DeterminantalPP}, Vendi Score~\cite{Friedman2022TheVS}, and IntDiv~\cite{Benhenda2017ChemGANCF} (a metric based on average pairwise dissimilarity, a concept originating in~\cite{Clarke2008NoveltyAD}) have emerged. Despite these advances, selecting appropriate unlabeled data remains challenging due to domain shifts and quality variations in real-world business documents~\cite{nigam2021document}, a gap our QUEST framework specifically addresses through its quality-aware approach.   
\section{Proposed Method}  

\subsection{System Overview}

\begin{figure}[h]
\centering
\includegraphics[width=0.99\linewidth]{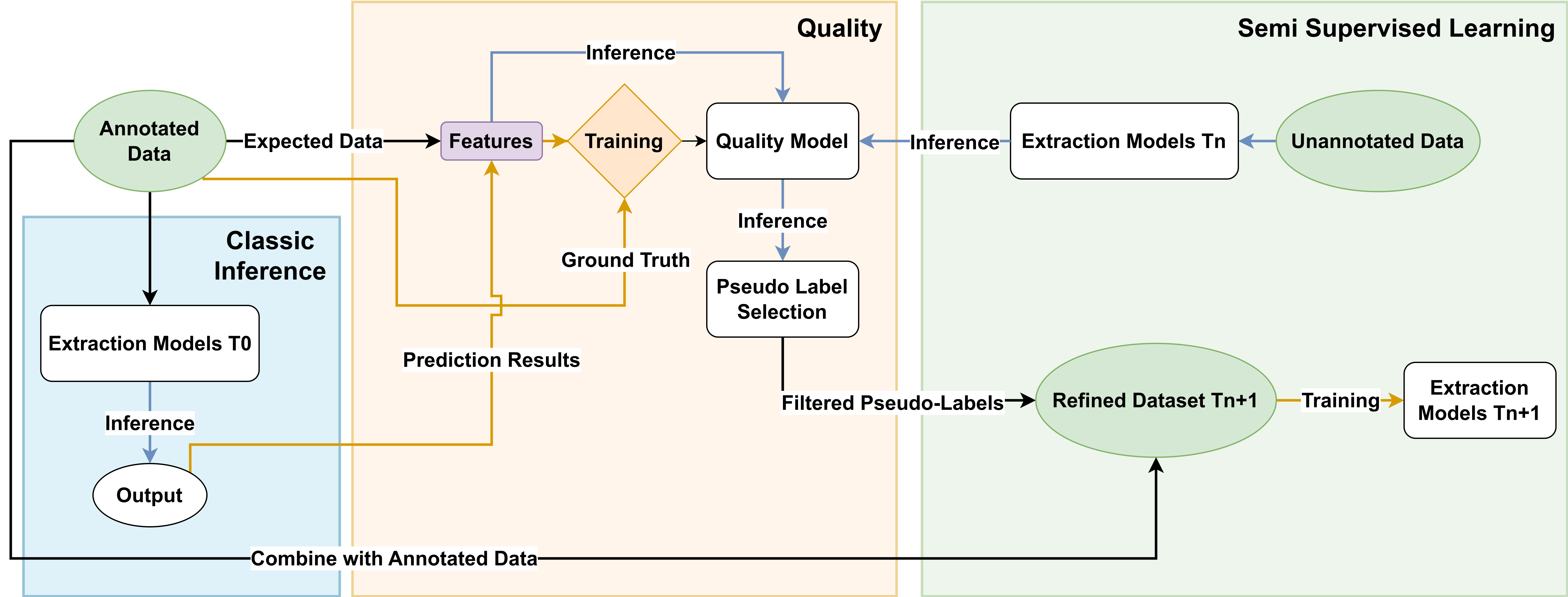}
\caption{Pipeline of our quality-aware SSL framework for table extraction.}
\label{fig:pipeline}
\end{figure}

As illustrated in Figure \ref{fig:pipeline}, our framework iteratively improves table extraction through quality-aware pseudo-labeling. The process begins with Classic Inference, where initial extraction models perform Table Detection (TD) followed by Table Structure Recognition (TSR) on annotated data. The annotated dataset is split 70/15/15 into training, validation, and test sets, maintained consistently across our framework.  
For prediction quality assessment, we employ a quality model trained on features from both ground truth (GT) and prediction results. This model evaluates tables using explainable characteristics, comparing documents against expected feature distributions. The Quality section of Figure \ref{fig:pipeline} illustrates this model's role in pseudo-label selection. The Semi-Supervised Learning phase applies trained extraction models to unannotated data. After selecting high-quality pseudo-labels, we implement diversity optimization to mitigate confirmation bias. From quality-filtered candidates, we select a subset maximizing diversity using DPP, Vendi Score, and IntDiv, ensuring a more representative training set. This process operates iteratively: as shown in the rightmost section of Figure \ref{fig:pipeline}, extraction models are trained from scratch at each step (Tn+1) using initial annotations combined with filtered pseudo-labels from previous iterations. The framework progressively refines extraction performance by leveraging both annotated and unannotated data through diverse, high-quality pseudo-labels.  

\subsection{Quality Assessment Model}

We introduce a quality assessment model that predicts F1 scores for extracted tables through a dual-phase approach: extracting domain-specific features and enriching them via statistical transformations. This design leverages expert knowledge while capturing data-driven patterns. The iterative data augmentation process can be formalized as:  

\begin{equation}  
    D_{t+1} = D_0 \cup f_t(U)  
\end{equation}  
where $D_0$ represents the initial annotated dataset, $U$ the unlabeled data pool, and $f_t$ the quality-aware selection function at iteration $t$. The dataset is curated to include only information that substantively benefits training, ensuring both quality and relevance. \\

\noindent\textbf{Feature Engineering}
% \subsubsection{Feature Engineering}
Our model uses layout, structural and contextual features of the table, as well as confidence scores from the extraction models: TD, TSR, and their product, TE, which reflects the sequential nature of the two steps. A detailed list of features is provided in Table~\ref{tab:features}.  \\

\noindent\textbf{Feature Transformation}
% \subsubsection{Feature Transformation}  
Let $x$ be the raw feature value, $\mu$ the mean, and $\sigma$ the standard deviation from the training distribution. Each base feature (21 in total) generates 5 engineered variants:  
\begin{itemize}  
    \item \textbf{Raw}: Original measurement  
    \item \textbf{Z-Score}: $(x - \mu)/\sigma$  
    \item \textbf{Deviation Magnitude}: $|x - \mu|/\sigma$  
    \item \textbf{Outlier Flag}: $1$ if $x > Q_{0.95}$, else $0$  
    \item \textbf{Normal Range}: $1$ if $\mu-\sigma \leq x \leq \mu+\sigma$, else $0$  
\end{itemize}  
This yields \(21 \times 5 = 105\) derived predictors, plus three confidence scores\\
(\(\text{conf}_{\text{TD}}, \text{conf}_{\text{TSR}}, \text{conf}_{\text{TE}}\)), totaling 108 features. This approach combines raw measurements with their training distribution positions, offering interpretable features rather than abstract embeddings.  

\begin{table}[htbp]  
\caption{Summary of quality assessment features}  
\label{tab:features}  
\centering  
\begin{tabular}{|p{0.35\textwidth}|p{0.65\textwidth}|}  
\hline  
\textbf{Feature Name} & \textbf{Description} \\
\hline  
Height Variation & Variation of the height of the rows \\
Width Variation & Variation of the width of the columns \\
Table Centering & Distance of the table from the image margins \\
Relative Position & Distance between table center and page center \\
Real Estate Usage & Proportion of the image area covered by the table \\
Content Isolation & Minimum distance to non-table elements (1.0 if none) \\
\hline  
Empty Cells Ratio & Ratio of empty cells in the table \\
Type Inconsistency & Column data type consistency using regex patterns \\
Row-to-Cell Ratio & Ratio of the number of rows to the total number of cells \\
Column-to-Cell Ratio & Number of columns divided by total cell count\\
Text Length Consistency & Maximum variation in cell text length per column \\
Alignment Inconsistency & Checks alignment of cells within extracted columns \\
Normalized Row Distances & Normalized distance between consecutive rows \\
Empty Cells Content Below & Ratio of empty cells with background text\\
\hline  
Content Continuity (in) & Ratio of aligned OCR text missed during extraction \\
Header Inside Suspicion & Ratio of header words not matching training examples \\
Header Outside Suspicion & Ratio of missed reference header words \\
Internal Whitespace Density & Text density between cells and TD bounds \\
Margin Whitespace Density & Text density in 20\% margin around TD \\
Content Type Transition & Content type continuity outside table via regex \\
Content Continuity (out) & Coordinate alignment continuity outside table\\
\hline  
\end{tabular}  
\end{table}

\noindent\textbf{Quality Model Training and Application}
% \subsubsection{Quality Model Training and Application}  
Our quality model is trained once on annotated data where GT enables accurate F1 calculation, using the same 70/15/15 data split as the extraction models. Let $Q: \mathcal{R}^{108} \rightarrow [0,1]$ represent our quality function that maps feature vectors to predicted F1 scores.  For a prediction $p$ on document $d$, we define its feature vector as:  
\begin{equation}  
X_p = [f_1, f_2, ..., f_{21}, \text{trans}(f_1), ..., \text{trans}(f_{21}), \text{conf}_{\text{TD}}, \text{conf}_{\text{TSR}}, \text{conf}_{\text{TE}}]  
\end{equation}  
Where $f_i$ are the base features from Table~\ref{tab:features} and $\text{trans}(f_i)$ represents the set of statistical transformations. The quality model is trained to minimize the error between predicted quality scores and actual F1 scores:  
\begin{equation}  
\min_{Q} \mathcal{L}(Q(X_p), F1_p)  
\end{equation}  
where $\mathcal{L}$ represents a suitable loss function, $X_p$ is the feature vector for prediction $p$, and $F1_p$ is the actual F1 score calculated by comparing prediction $p$ to its GT. This trained model remains unchanged throughout the SSL process, providing consistent quality estimation for reliable pseudo-label selection.  \\

%\noindent\textbf{Pseudo-label Selection}
\subsection{Pseudo-label Selection}  
Our pseudo-label selection process has three key steps: (1) gathering unlabeled data, (2) filtering out low-quality entries, and (3) selecting a maximally diverse subset for pseudo-labeling.  \\

\noindent\textbf{Unlabeled Data Curation}
% \subsubsection{Unlabeled Data Curation}  
We assume access to a large corpus of unlabeled images. To identify those that are most likely to contain useful tables, we apply the following filtering steps:  
\begin{enumerate}  
    \item \textbf{Empty Check:} Discard blank or near-blank images.  

    \item \textbf{Orientation Check:} Among non-empty images, retain only those recognized as upright via an OCR-based orientation detection.  
    \item \textbf{Table Presence Verification:}  
    \begin{enumerate}  
        \item Run a table-detection model trained on our annotated data. If a table is detected, keep the image.  
        \item Otherwise, apply a publicly available table-detection model with a confidence threshold \(\theta_{\text{high}}\). If the confidence is at least \(\theta_{\text{high}}\), keep the image.  
        \item If the confidence is below \(\theta_{\text{high}}\) but above another threshold \(\theta_{\text{mid}}\), we query a vision-language model (VLM) with a prompt such as: “Is there a table of items in the image? Respond only with True or False.” If the response is “True,” keep the image; otherwise, discard it.  
    \end{enumerate}  
\end{enumerate}  

\noindent\textbf{Quality Filtering}
% \subsubsection{Quality Filtering}  
Each extracted table receives a quality score (estimated F1) from model $Q$. We retain tables where:   
\begin{equation}  
Q(X_p) \geq \alpha,  
\end{equation}  
with threshold $\alpha$ and feature vector $X_p$ as defined in the quality model section. This strategy enhances traditional confidence-based filtering by incorporating multiple quality indicators. \\

\noindent\textbf{Diversity Selection}
% \subsubsection{Diversity Selection}  
From the quality-filtered set \(\mathcal{D}_q\), we seek a subset that is both diverse and informative. Our approach comprises the following steps:  

\begin{enumerate}  
    \item \textbf{DPP-Based Subset Selection.}  
    We construct an RBF kernel:  
    \begin{equation}  
    K_{ij}   
    = \exp\!\left(-\frac{\|x_i - x_j\|^2}{2\sigma^2}\right),  
    \quad  
    \sigma = \text{median}\{\|x_i - x_j\|\},  
    \end{equation}  
    and select a subset \(S^*\) of size \(k\) that maximizes the submatrix determinant:   
    \begin{equation}  
    S^* = \arg\max_{S \subset \mathcal{D}_q,\, |S|=k} \det(K_S).  
    \end{equation}  

    \item \textbf{Diversity Measures.}  
    We quantify the diversity of a set \(T\) using two main metrics:  
    \begin{enumerate}  
        \item \textit{Vendi Score (VS).}  
        \begin{equation}  
        \operatorname{VS}(T) = \exp\!\biggl(-\sum_{i=1}^{n}\lambda_i \log \lambda_i\biggr),  
        \end{equation}  
        where \(\{\lambda_i\}\) are the eigenvalues of the normalized kernel \(K / n\). This can be interpreted as the effective number of unique elements in \(T\).  

        \item \textit{Internal Diversity (IntDiv).}  
        \begin{equation}  
        \operatorname{IntDiv}(T)   
        = 1 - \frac{1}{n(n-1)} \sum_{i \neq j}   
        \exp\!\left(-\frac{\|x_i - x_j\|^2}{2\sigma^2}\right).  
        \end{equation}  
        This measures average pairwise dissimilarity within \(T\).  
    \end{enumerate}  

    \item \textbf{Candidate Subset Evaluation.}  
    Let \(A_{\text{train}}\) be the annotated training set, and \(S_k\) a DPP-sampled subset of size \(k\). We form:  
    \begin{equation}  
    T_k = A_{\text{train}} \cup S_k  
    \end{equation}  
    and define its overall diversity as:  
    \begin{equation}  
    D(T_k) = \operatorname{VS}(T_k)\,\times\,\operatorname{IntDiv}(T_k).  
    \end{equation}  
    The optimal subset \(S_{\text{opt}}\) maximizes:  
    \begin{equation}  
    S_{\text{opt}}  
    = \arg\max_{S_k}  
    D\!\bigl(A_{\text{train}} \cup S_k\bigr).  
    \end{equation}  
\end{enumerate}

\section{Experiments}  

\subsection{Experimental Setup}  

\noindent\textbf{Datasets}
% \subsubsection{Datasets}  
We evaluate on two datasets with a 75/15/15 train/val/test split: (1) A private business document collection with 1,639 annotated tables and 10,109 unlabeled documents, primarily invoices and financial statements containing structured tables; (2) The public DocILE dataset~\cite{vSimsa2023DocILEBF}, processed following Thomas et al.~\cite{thomas2025raptorrefinedapproachproduct}, with 954 annotated tables and 19,196 unlabeled multi-page PDFs from batch one, spanning diverse business layouts\footnote{\url{https://github.com/eliottthomas99/Data_QUEST/tree/main}}.  \\

\noindent\textbf{Implementation details}
% \subsubsection{Implementation Details}  
We select YOLOv9-T~\cite{Wang2024YOLOv9LW} as our extraction model for its lightweight architecture and stability at the time of experiments. Training from scratch uses SGD (lr=0.01, batch=16) for up to 500 epochs with early stopping (patience=100). Quality assessment uses XGBoost, with parameters tuned via randomized search: estimators (100-1000), max depth (3-12), and learning rate (0.01-0.3). For unlabeled data filtering, we employ three TD models: our YOLOv9, DETR fine-tuned on ICDAR19\footnote{\url{https://huggingface.co/TahaDouaji/detr-doc-table-detection}}~\cite{Gao2019ICDAR2C}, and Qwen2-VL-7B~\cite{Qwen-VL,Qwen2VL}, using thresholds $\theta_{\text{high}} = 0.95$ and $\theta_{\text{med}} = 0.5$. Quality threshold $\alpha$ is set to 0.9, balancing high confidence with achievability for business documents.\\

\noindent\textbf{Evaluation Protocol}
% \subsubsection{Evaluation Protocol}  
We evaluate using GRITS-CON~\cite{Smock2022GriTSGT} for structural and textual accuracy, plus:  
\begin{itemize}  
    \item Empty prediction rate: percentage of documents with no detected tables  
    \item Document-level analysis: F1-score changes between iterations (per doc)  
\end{itemize} 
Our evaluation examines: (1) if SSL with pseudo-labels outperforms training on labeled data alone, and (2) how our quality-based selection compares to traditional confidence filtering, using each method's best model.  

\subsection{Results}  

\begin{figure}[h]
\centering
\includegraphics[width=0.99\linewidth]{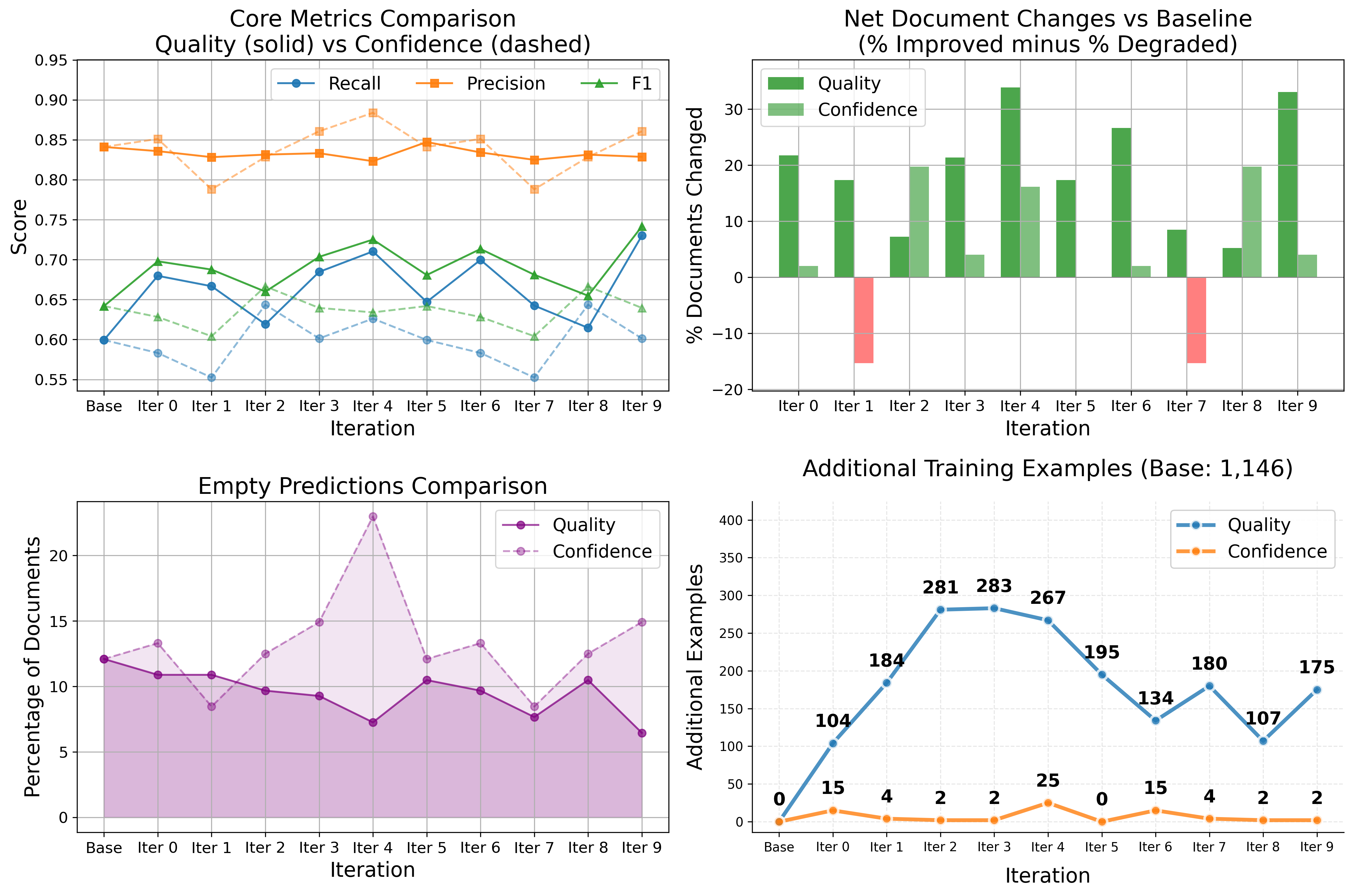}
\caption{SSL Performance Metrics on Private Business Dataset: Quality-based vs Confidence-based Approaches}  
\label{fig:ssl_figure_private}
\end{figure}

\begin{figure}[h]
\centering
\includegraphics[width=0.99\linewidth]{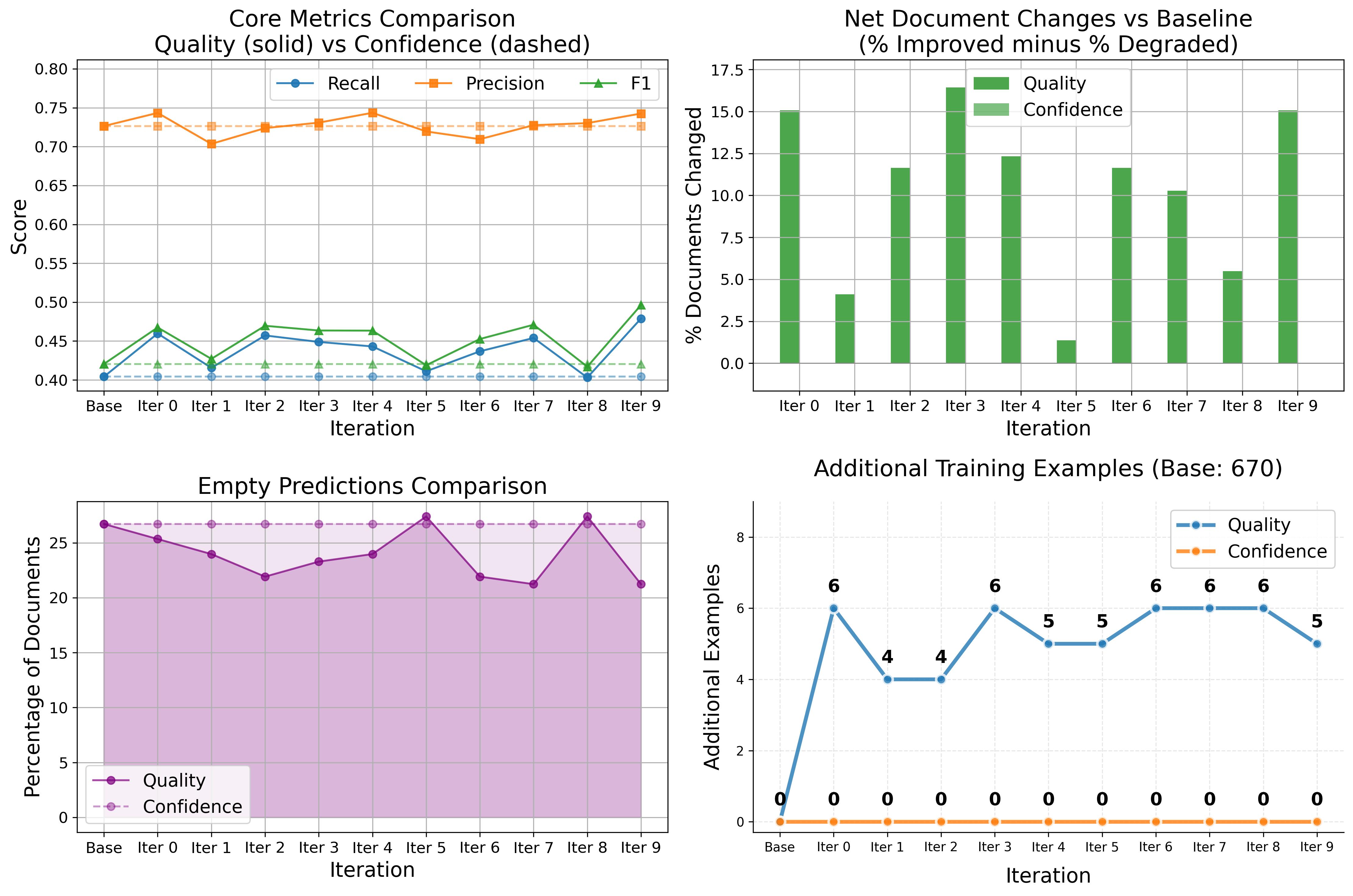}
\caption{SSL Performance Metrics on DocILE Dataset: Quality-based vs Confidence-based Approaches}  
\label{fig:ssl_figure_docile}
\end{figure}

\noindent\textbf{Performance Analysis}
% \subsubsection{Performance Analysis}  
Our experimental results demonstrate the superior performance of the quality-aware approach compared to the confidence-based method across both datasets, as shown in Figure \ref{fig:ssl_figure_private} and \ref{fig:ssl_figure_docile}. The effectiveness of our approach varies significantly between datasets, primarily due to their inherent characteristics and initial baseline performances.

On the private dataset, where baseline performance starts at an F1 score of 64\%, both SSL approaches show improvement, but with notably different patterns. The quality-based approach achieves the best performance with an F1 score of 74\%, while maintaining precision at 83\% despite reducing empty predictions from 12\% to 6.5\% (Table \ref{tab:best_res}). This reduction in empty predictions is particularly significant as empty predictions artificially inflate precision (resulting in Recall=0, Precision=1, F1=0). The confidence-based approach, while achieving its best results at iteration 2, begins to loop after iteration 5, limiting its potential for further enhancement. The net document change is defined as the difference between the percentage of documents that improved and degraded in F1 score. In our case, this +33\% net improvement demonstrates the robustness of the quality-based approach, as it indicates that significantly more documents benefited from the method than were negatively impacted by it.

The DocILE dataset presents a more challenging scenario, starting with a lower baseline F1 score of 42\% (Table \ref{tab:best_res}). In this context, we observe a stark contrast between approaches: the confidence-based method fails entirely, generating no pseudo-labels above the quality threshold and remaining at baseline performance. In contrast, our quality-oriented approach successfully identifies valuable candidates for training. This demonstrates two crucial points: first, SSL methods require adequate initial performance to generate useful pseudo-labels, and second, our method's ability to identify high-quality samples proves especially valuable in challenging conditions. The effectiveness of our approach is reflected in a net document improvement of 15\%, reaching an F1 score of 50\% while reducing empty predictions from 27\% to 22\%.

These results demonstrate that our approach can work effectively across a variety of business documents, suggesting potential adaptability to other document types. The DocILE dataset's more modest improvements can be attributed to three main factors: its greater document variety, smaller training set (670 vs 1,146 documents), and its original focus on line item retrieval rather than complete table extraction. This distinction is significant as line item retrieval only annotates specific columns of interest within tables, while table extraction requires all table content to be annotated. This partial annotation scheme makes it more challenging for models to learn complete table structures, as they must simultaneously identify table boundaries while implicitly learning which columns are relevant. Despite these challenges, our quality-aware approach demonstrates consistent improvement across both datasets.  

\begin{table}  
\caption{Performance Comparison of SSL Approaches on Both Datasets}  

\label{tab:best_res}  
\centering  
\begin{tabular}{llccccccc}  
\hline  
Dataset & Approach & P (\%) & R (\%) & F1 (\%) & \#Docs & Rel. Imp. & Empty (\%) & Iter\\
\hline  
\multirow{3}{*}{Private}   
        & Baseline    & \textbf{84} & 60 & 64 & 1,146 & - & 12 & - \\
        & Confidence  & 83 & 65 & 67 & 1,148 & +20\% & 10 & 2 \\
        & Quality     & 83 & \textbf{73} & \textbf{74} & \textbf{1,321} & \textbf{+33\%} & \textbf{6.5} & 9 \\
\hline  
\multirow{3}{*}{DocILE}  
        & Baseline    & 73 & 40 & 42 & 670 & - & 27 & - \\
        & Confidence  & 73 & 40 & 42 & 670 & - & 27 & - \\
        & Quality     & \textbf{74} & \textbf{48} & \textbf{50} & \textbf{675} & \textbf{+15\%} & \textbf{22} & 9 \\
\hline  
\end{tabular}  
\end{table}  

\noindent\textbf{Quality Model Evaluation}
% \subsubsection{Quality Model Evaluation}  
Our quality assessment framework demonstrates significant improvements over traditional confidence scores through three key findings: stronger correlation with actual performance, dataset-specific feature importance patterns, and the complementary role of confidence metrics. The framework achieves substantially stronger correlation with F1 scores than confidence metrics alone. On our private dataset (Figure~\ref{fig:teaser}), it reaches r=0.80 (RMSE=0.13) versus confidence scores' r=0.22 (RMSE=0.24). For DocILE, the quality model maintains r=0.67 (RMSE=0.22) compared to confidence scores' r=0.46 (RMSE=0.26). Three factors explain this superiority:  

\begin{figure}[h]  
\centering  
\includegraphics[width=0.99\linewidth]{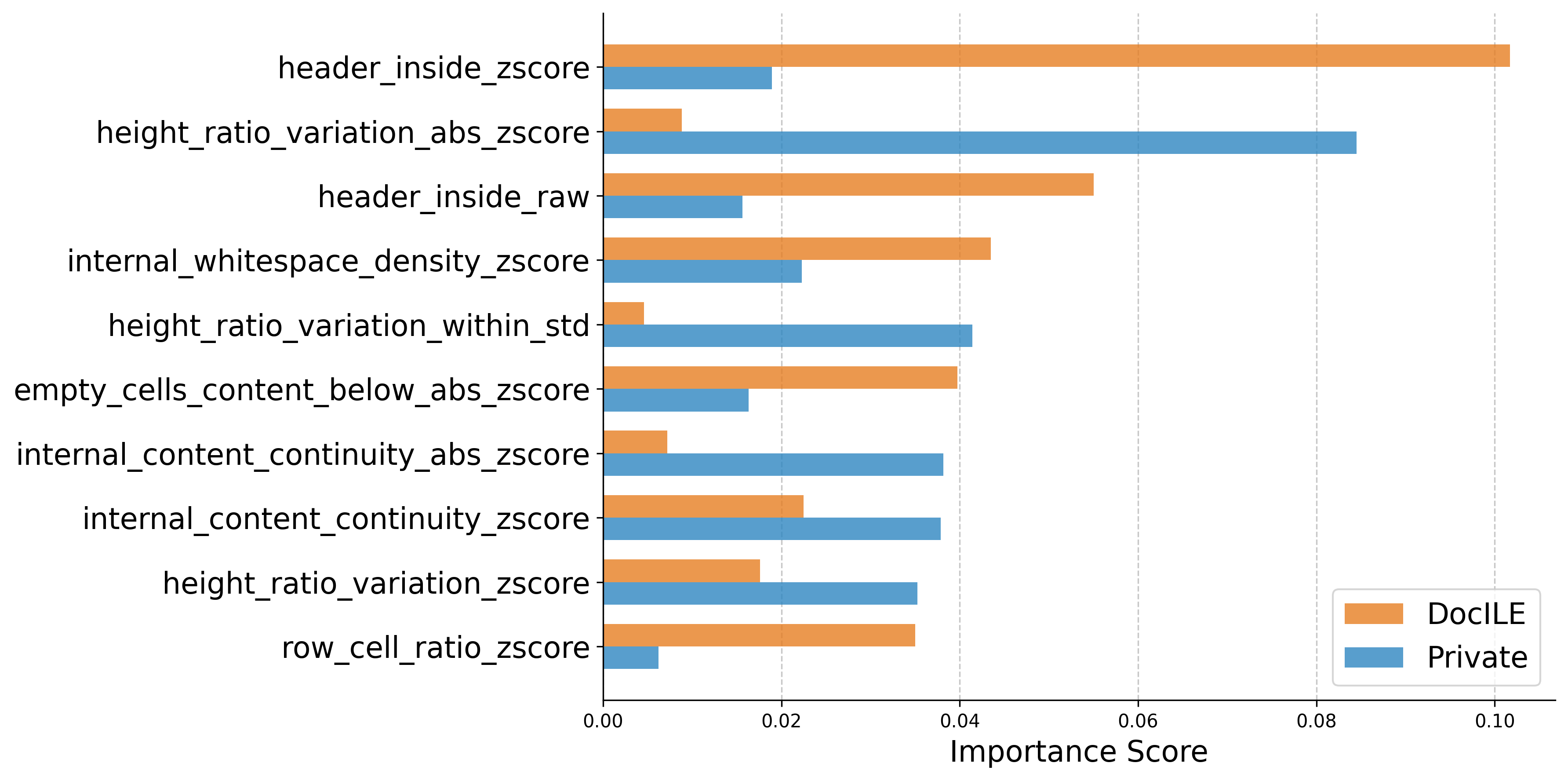}  
\caption{Feature importance patterns in quality estimation across datasets}  
\label{fig:feature_importance_xgb}  
\end{figure} 

First, statistical transformations dominate feature importance (Figure~\ref{fig:feature_importance_xgb}), particularly z-score variants occupying 8 of the top 10 positions. These transformations compare new table features against statistics from known correct tables in our training set. Regular z-scores (5 features) indicate whether measurements deviate above or below expected values, while absolute z-scores (3 features) capture the magnitude of these deviations regardless of direction.

Second, dataset characteristics shape feature priorities. Private documents, typically containing invoice-style tables with regular row patterns, rely on layout features like \texttt{height\allowbreak\_ratio\allowbreak\_variation\allowbreak\_abs\allowbreak\_zscore} ($\sim$0.085) to detect parsing errors through unexpected height variations. DocILE's complex documents, containing multiple table-like elements, prioritize structural features such as \texttt{header\allowbreak\_inside\allowbreak\_zscore} ($\sim$0.10) to verify whether header content matches expected patterns. Contextual features like \texttt{internal\allowbreak\_whitespace\allowbreak\_density} remain consistently important across both datasets.

Third, confidence metrics show consistently low importance across datasets, with none in the top 25 features. The highest-ranked confidence metrics are conf\_TD (31st, private dataset) and conf\_TE (28th, DocILE). This pattern holds across all confidence types (TD, TSR, TE). The finding validates our feature set: while quality indicators vary by dataset, our structural and statistical features better capture table quality than raw confidence scores.

This combination of statistical normalization, dataset-aware prioritization, and contextual analysis enables robust quality estimation across document types, as evidenced by the strong correlations in Figure~\ref{fig:teaser}.

\section{Discussion}  

\subsection{Qualitative Analysis}
\begin{figure}[h]  
\centering  
\begin{subfigure}[t]{0.32\textwidth}  
\centering  
\includegraphics[width=\textwidth]{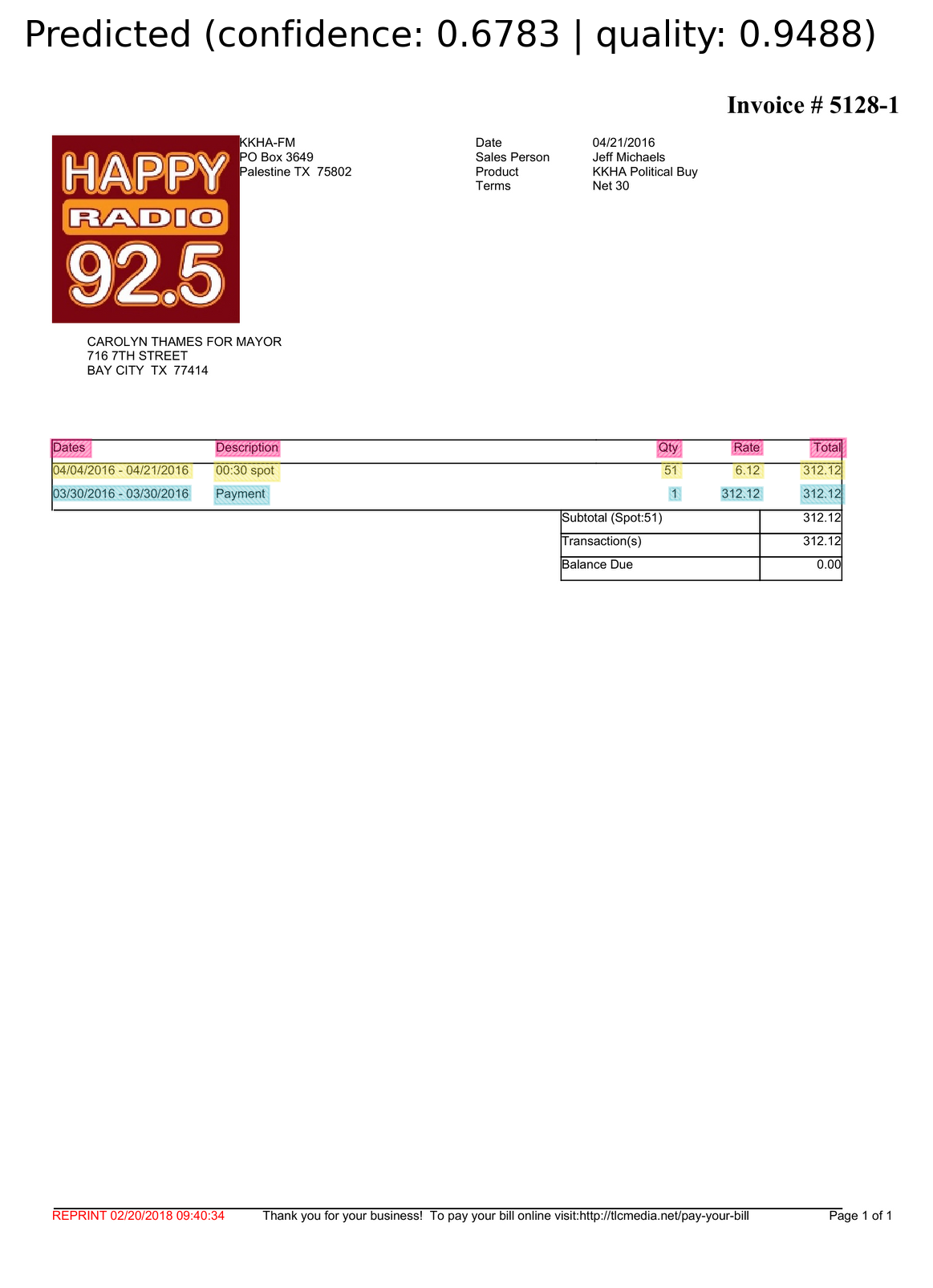}  
\caption{Good doc: Q:k C:r}  
\label{fig:case1}  
\end{subfigure}  
\begin{subfigure}[t]{0.32\textwidth}  
\centering  
\includegraphics[width=\textwidth]{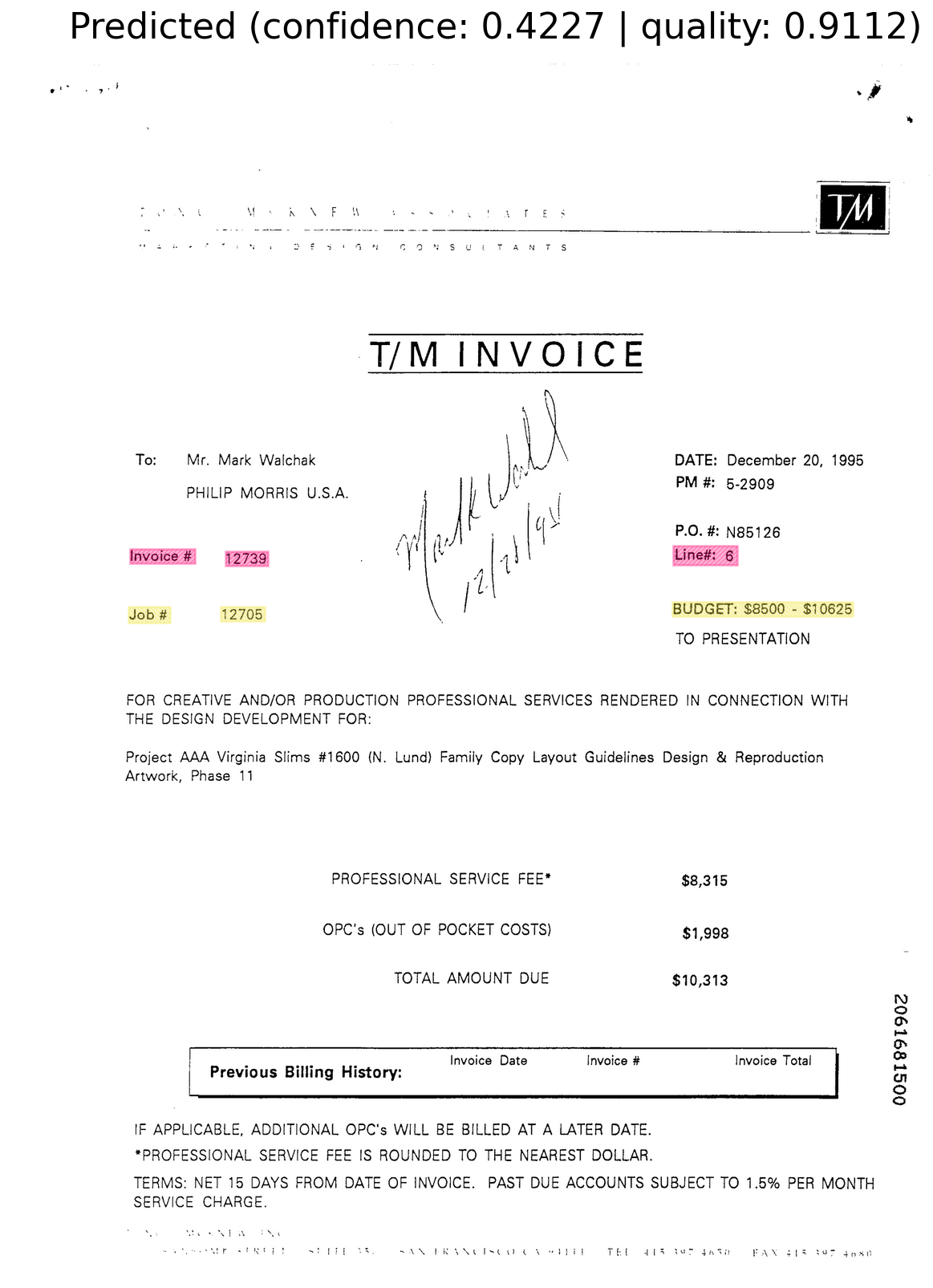}  
\caption{Bad doc: Q:k C:r}  
\label{fig:case2}  
\end{subfigure}  
\begin{subfigure}[t]{0.32\textwidth}  
\centering  
\includegraphics[width=\textwidth]{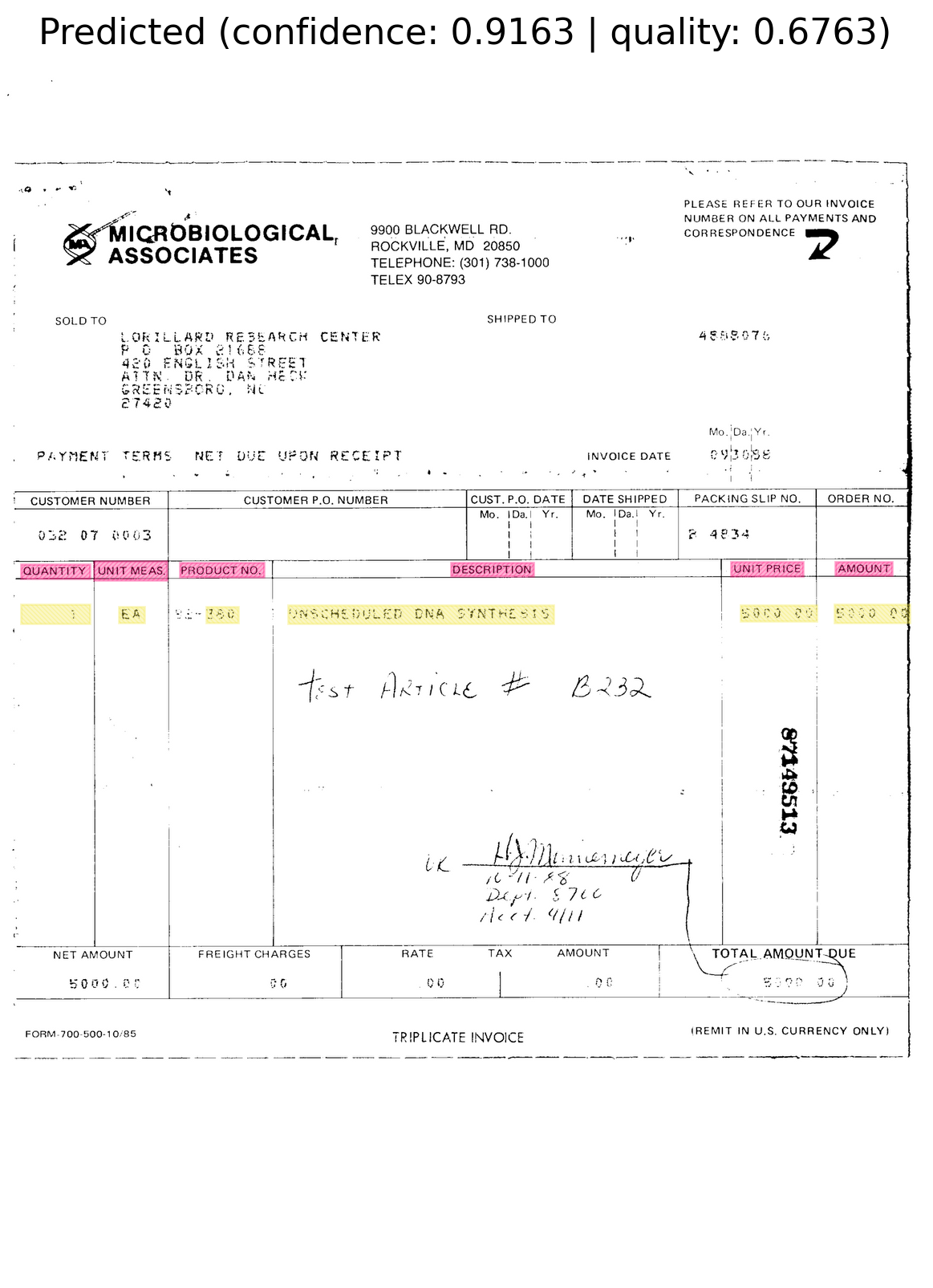}  
\caption{Good doc: Q:r C:k}  
\label{fig:case3}  
\end{subfigure}  
\caption{Quality (Q) vs confidence (C) divergence cases, where k=keep, r=remove.}  
\label{fig:qualitative_analysis}  
\end{figure}

\noindent\textbf{Model Disagreements}
% \subsubsection{Model Disagreements}  
To complement our quantitative evaluation, we analyze three representative cases from the DocILE dataset where quality and confidence models diverge (Figure \ref{fig:qualitative_analysis}). The most common scenario (Figure \ref{fig:case1}) shows the quality model correctly accepting a good document (quality: 0.95) that confidence rejects (0.68). A rarer case (Figure \ref{fig:case2}) shows confidence correctly rejecting a poor document (0.42) that quality accepts (0.91). Finally, an extremely rare case (Figure \ref{fig:case3}, observed once) shows quality incorrectly rejecting a good document (0.67) that confidence accepts (0.91). This document was excluded by diversity checks. These examples highlight the quality model's ability to capture document quality while exposing its occasional limitations.   

\begin{figure}[h]  
\centering  
\begin{subfigure}[t]{0.48\textwidth}  
\centering  
\includegraphics[width=\textwidth]{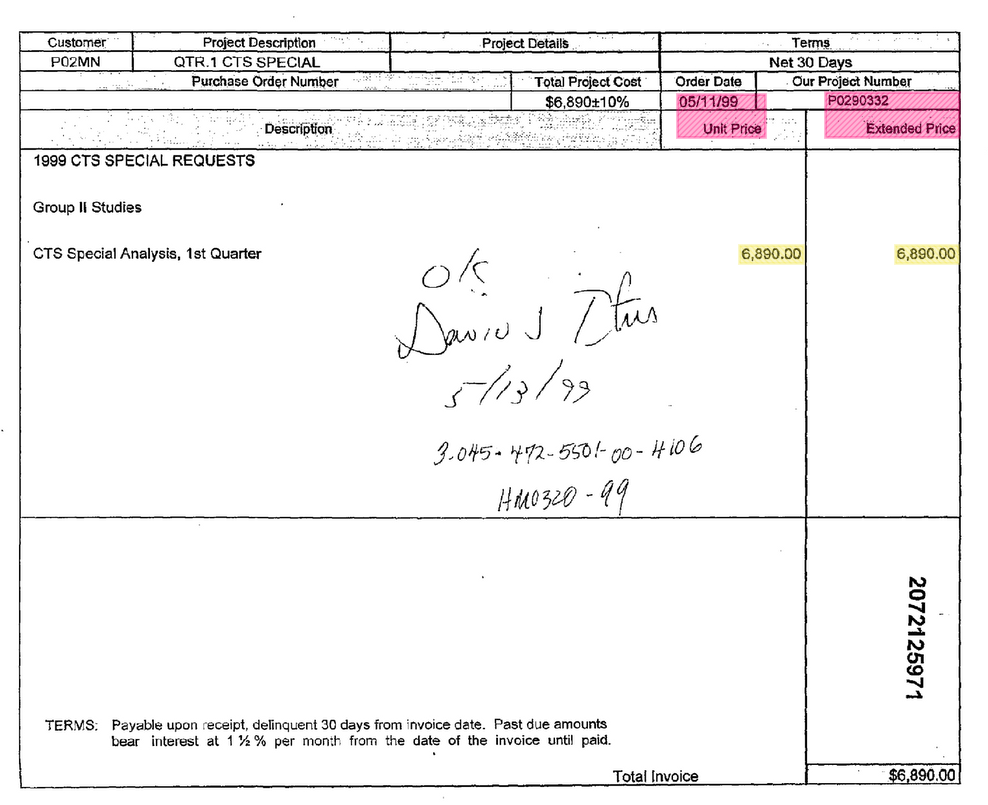}  
\caption{Base model: Missing column detection}  
\label{fig:improve1_before}  
\end{subfigure}  
\begin{subfigure}[t]{0.48\textwidth}  
\centering  
\includegraphics[width=\textwidth]{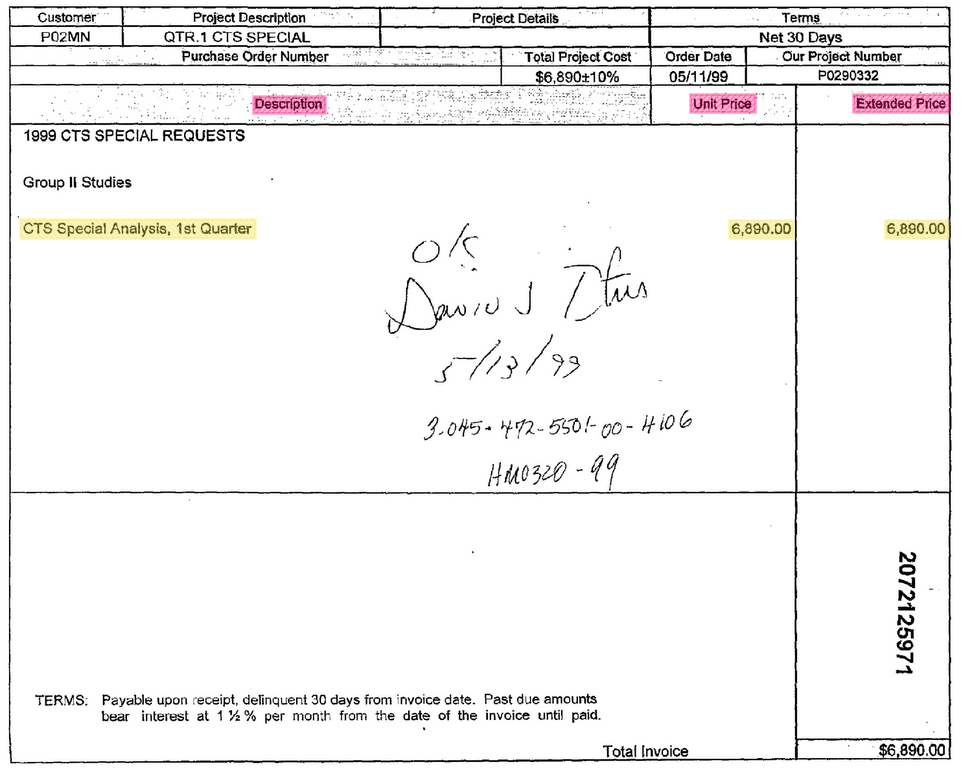}  
\caption{Improved model: Complete detection}  
\label{fig:improve1_after}  
\end{subfigure}  

\vspace{2mm}  

\begin{subfigure}[t]{0.48\textwidth}  
\centering  
\includegraphics[width=\textwidth]{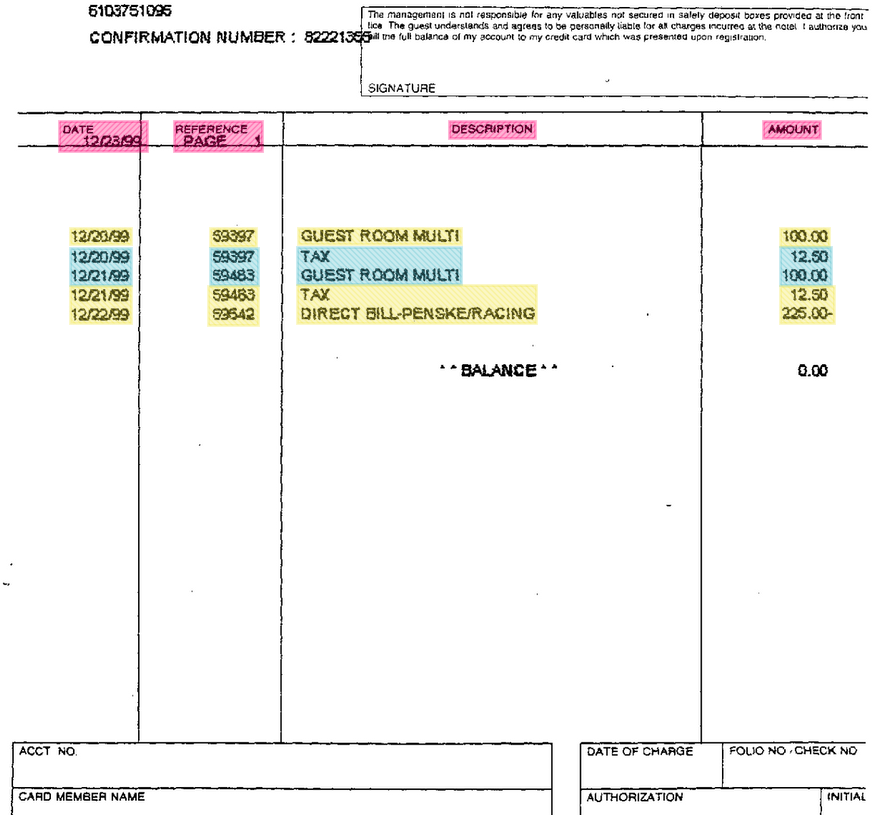}  
\caption{Base model: Row merging}  
\label{fig:improve2_before}  
\end{subfigure}  
\begin{subfigure}[t]{0.48\textwidth}  
\centering  
\includegraphics[width=\textwidth]{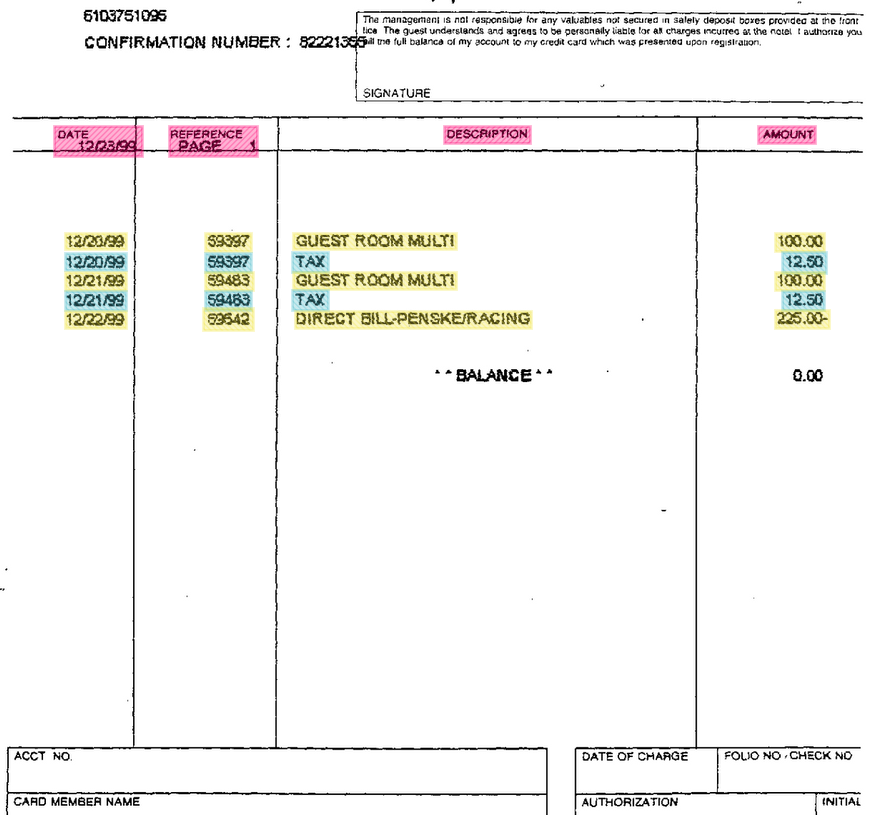}  
\caption{Improved model: Correct separation}  
\label{fig:improve2_after}  
\end{subfigure}  
\caption{Base model vs iteration 9 improvements (DocILE).}  
\label{fig:quality_improvements}  
\end{figure}

\noindent\textbf{Detection Improvements}
% \subsubsection{Detection Improvements}  
We analyze two DocILE documents in which our improved model (iteration 9) corrects the errors of the base model (Figure \ref{fig:quality_improvements}). The first case shows successful detection of a previously missed column, improving F1 from 0.4 to 1.0 (Figures \ref{fig:improve1_before}, \ref{fig:improve1_after}). The second demonstrates correct row separation instead of erroneous merging, raising F1 from 0.2 to 0.75 (Figures \ref{fig:improve2_before}, \ref{fig:improve2_after}).

\subsection{Limitations}  

\noindent\textbf{Computational Considerations}  
Our SSL pipeline requires ~15 hours per iteration on an NVIDIA RTX A6000 GPU with Intel Xeon Silver 4210R CPU. For our private dataset (1,146 documents), this comprises 12.5h training and 2.8h pseudo-label generation; DocILE requires 10h/4.5h due to its larger unlabeled set. While industrially feasible, we limited testing to 10 iterations. Future work could reduce costs via progressive learning, model distillation, or selective retraining to enable more iterations.  

\noindent\textbf{Initial Model Requirements}  
QUEST requires initial models capable of quality predictions, as DocILE results show: only 4-6 predictions per iteration met quality thresholds from thousands of candidates, while our private dataset's stronger baseline yielded more viable pseudo-labels. Though restrictive, this strict filtering ensures reliable improvement. Future work could explore dynamic thresholds based on dataset characteristics.  

\noindent\textbf{Technical Constraints}  
The framework currently processes single tables without spanning cells, with multi-table handling planned. Quality assessment performance depends on training examples, with DocILE results indicating room for improvement in low-resource settings.  

\subsection{Industrial Applications}  

\noindent\textbf{Quality Assessment as Rejection Mechanism}  
Our quality model delivers interpretable predictions (RMSE: 0.13 between predicted/actual F1 scores) to automatically reject low-quality extractions, identifying issues like abnormal spacing, missed content, and structural inconsistencies. This reduces manual verification needs while offering more reliable metrics than confidence scores.  \\ 

\noindent\textbf{Domain Adaptability}  
QUEST particularly benefits TSR where structural elements must form a coherent whole. While adapting to new domains requires feature engineering based on domain-specific patterns, our interpretable approach offers advantages over automated feature extraction through embeddings. The framework scales effectively with multi-information processing tasks.  \\

\noindent\textbf{Deployment Considerations}  
QUEST's modular design enables flexible deployment with adjustable quality thresholds (recommended: 0.9), benefiting from higher thresholds with stronger models. The framework supports targeted human-in-the-loop integration for low-quality and borderline-quality predictions, optimizing effort while maintaining automation.    

\section{Conclusion}

This paper introduced QUEST, a quality-aware semi-supervised learning framework for table extraction that significantly improves F1 scores (from 64\% to 74\% on our private dataset and from 42\% to 50\% on DocILE) using unlabeled business documents. Our key innovation replaces confidence scores with an interpretable quality assessment model, enabling robust pseudo-label selection while mitigating confirmation bias. This approach effectively leverages companies' existing unlabeled document repositories, with gains primarily in recall. Future work includes handling complex tables and reducing annotation requirements.  

\section{Acknowledgments}

This research was suported by the CIFRE PhD program funded by the ANRT, by the French government through the France Relance program, by the ANR agency under grant number ANR-21-PRRD-0010-01 (LabCom Ideas), as well as by the YOOZ company.

% \begin{credits}  
% \subsubsection{\ackname}   
% This work was conducted as part of the CIFRE PhD program of Eliott THOMAS, established between Yooz company and the L3i Laboratory at La Rochelle Université. The authors would like to thank the French National Association for Research and Technology (ANRT) for funding this CIFRE PhD.  

% \subsubsection{\discintname}  
% The authors declare no competing interests that are relevant to the content of this article. Authors affiliated with Yooz are employees of the company, which is disclosed in the author affiliations. This industrial-academic collaboration is transparently represented in the authorship and does not influence the scientific integrity of the research presented.  
% \end{credits}  
\bibliographystyle{splncs04}
\bibliography{mybibliography}

\newpage
\appendix
\section{Overview}  
While the main paper is self-contained and presents all essential contributions and results, this supplementary document provides additional details for interested readers. It includes extended implementation specifications, complementary experimental results, and in-depth analyses that could offer further insights into our approach.

\section{Implementation Details}  

\subsection{Model Hyperparameters}  

\subsubsection{Table Extraction Model (YOLOv9-T)}  
We use YOLOv9-T as our primary extraction model, with training configuration detailed in Table~\ref{tab:yolo_params}.  

\begin{table}[h]  
\caption{YOLOv9-T Training Parameters}  
\label{tab:yolo_params}  
\centering  
\small  
\begin{tabular}{llllll}  
\toprule  
\multicolumn{2}{c}{\textbf{Optimization}} & \multicolumn{2}{c}{\textbf{Training \& Loss}} & \multicolumn{2}{c}{\textbf{Augmentation}} \\
\cmidrule(lr){1-2} \cmidrule(lr){3-4} \cmidrule(lr){5-6}  
Parameter & Value & Parameter & Value & Parameter & Value \\
\midrule  
Initial LR & 0.01 & Max epochs & 500 & HSV-Hue & ±0.015 \\
Final LR & 0.0001 & Early stop patience & 100 & HSV-Saturation & ±0.7 \\
Momentum & 0.937 & Warmup epochs & 3.0 & HSV-Value & ±0.4 \\
Weight decay & 0.0005 & Box loss gain & 7.5 & Translation & ±0.1 \\
 &  & Class loss gain & 0.5 & Scale & 0.9 \\
 &  & Object loss gain & 0.7 & Horiz. flip prob. & 0.5 \\
 &  & DFL loss gain & 1.5 &  &  \\
 &  & IoU threshold & 0.20 &  &  \\
\bottomrule  
\end{tabular}  
\end{table}  

\subsubsection{Quality Assessment Model (XGBoost)}  
The quality assessment model uses XGBoost with hyperparameters optimized through randomized search cross-validation. Table~\ref{tab:xgb_params} details the search space and final configuration.  

\begin{table}[h]  
\caption{XGBoost Quality Assessment Model Configuration}  
\label{tab:xgb_params}  
\centering  
\begin{tabular}{lll}  
\toprule  
\textbf{Parameter} & \textbf{Search Space} & \textbf{Description} \\
\midrule  
n\_estimators & [100, 1000] & Number of boosting rounds \\
max\_depth & [3, 12] & Maximum tree depth \\
learning\_rate & [0.01, 0.3] & Step size shrinkage \\
min\_child\_weight & [1, 7] & Minimum sum of instance weight \\
subsample & [0.6, 1.0] & Subsample ratio of training instances \\
colsample\_bytree & [0.6, 1.0] & Subsample ratio of columns \\
\midrule  
\multicolumn{3}{l}{\textit{Search Configuration:}} \\
\multicolumn{3}{l}{- Random search with 100 iterations} \\
\multicolumn{3}{l}{- 5-fold cross-validation} \\
\multicolumn{3}{l}{- Optimization metric: R² score} \\
\bottomrule  
\end{tabular}  
\end{table}

\subsection{Quality Assessment Features}  

Our quality assessment model evaluates table extraction results through a comprehensive set of features designed to capture structural, content, and contextual aspects. Before describing the features, we define several key concepts:  

\begin{itemize}  
    \item \textbf{Header Words}: We identified common header words from the training set, retaining those appearing in at least 10 different documents. This threshold captures approximately 90\% of header terminology while filtering noise.  
    
    \item \textbf{Content Types}: We use regular expression patterns to classify cell content into categories including numerical, date-like, amount-like, and text.  
    
    \item \textbf{Cell Alignment}: Text alignment within cells is classified as left, right, center, or unknown based on whitespace distribution.  
\end{itemize}  

The features are organized into three categories: structural features, content features, and contextual features.  

\subsubsection{Structural Features}  
These features assess the physical attributes and positioning of the extracted table:  

\begin{description}  
    \item[Height Variation] Measures the standard deviation of row heights, indicating regularity of structure.  
    
    \item[Width Variation] Measures the standard deviation of column widths, indicating regularity of structure.  
    
    \item[Table Centering] Calculates the distance between table edges and image margins, normalized by page dimensions.  
    
    \item[Relative Position] Measures the Euclidean distance between the table center and page center, normalized by page diagonal.  
    
    \item[Real Estate Usage] Calculates the proportion of the document area covered by the table.  
    
    \item[Content Isolation] Returns the minimum distance to non-table elements (1.0 if none exist nearby), providing insight into table separation.  
\end{description}  

\subsubsection{Content Features}  
These features evaluate the internal consistency and patterns within the extracted table:  

\begin{description}  
    \item[Empty Cells Ratio] Calculates the proportion of extracted cells that contain no text.  
    
    \item[Type Inconsistency] For each column, we identify the dominant content type using regex patterns and measure the proportion of cells deviating from this type. The feature returns the maximum inconsistency across all columns.  
    
    \item[Row-to-Cell Ratio] Returns the number of rows divided by the total cell count, helping identify merged or split cells.  
    
    \item[Column-to-Cell Ratio] Returns the number of columns divided by the total cell count, complementing row-to-cell ratio.  
    
    \item[Text Length Consistency] For each column, computes the standard deviation of text length (character count). Returns the maximum standard deviation across columns, normalized by mean length.  
    
    \item[Alignment Inconsistency] Calculates the proportion of columns containing cells with inconsistent text alignment.  
    
    \item[Normalized Row Distances] Computes the standard deviation of distances between consecutive rows, normalized by table height.  
    
    \item[Empty Cells Content Below] Measures the proportion of empty cell area that overlaps with underlying OCR text. For example, if there are two empty cells of sizes 10 and 30 square units, and the 30-unit cell has 10 units overlapping with OCR text, the ratio would be 10/40 = 0.25.  
\end{description}  

\subsubsection{Contextual Features}  
These features assess the relationship between the extracted table and surrounding document elements:  

\begin{description}  
    \item[Internal Content Continuity] Identifies OCR text within table boundaries that was not captured in the extracted cells. Returns the ratio of area covered by these "missed" text elements to the total text area within the table.  
    
    \item[Header Inside Suspicion] Calculates the proportion of words in extracted headers that do not appear in our training-derived list of common header words.  
    
    \item[Header Outside Suspicion] Calculates the proportion of words in non-header regions that match our list of common header words, potentially indicating missed headers.  
    
    \item[Internal Whitespace Density] Measures text density in the area between the table detection boundary and the smallest rectangle containing all cells. Returns the proportion of this area covered by OCR text.  
    
    \item[Margin Whitespace Density] Calculates text density in a margin around the table (20\% of table dimensions), indicating potential context or related information.  
    
    \item[Content Type Transition] Evaluates whether content immediately outside table boundaries matches the patterns of adjacent table columns/rows, which could indicate missing data.  
    
    \item[External Content Continuity] Assesses alignment continuity of text elements immediately outside table boundaries with the table structure, helping identify potentially missed rows or columns.  
\end{description}  

These features collectively provide a comprehensive assessment of extraction quality, capturing not only the structural integrity of the table but also its semantic coherence and relationship with the document context.  

\section{Augmented Documents in DocILE}

\begin{figure}[ht]  
    \centering  
    \includegraphics[width=0.7\textwidth]{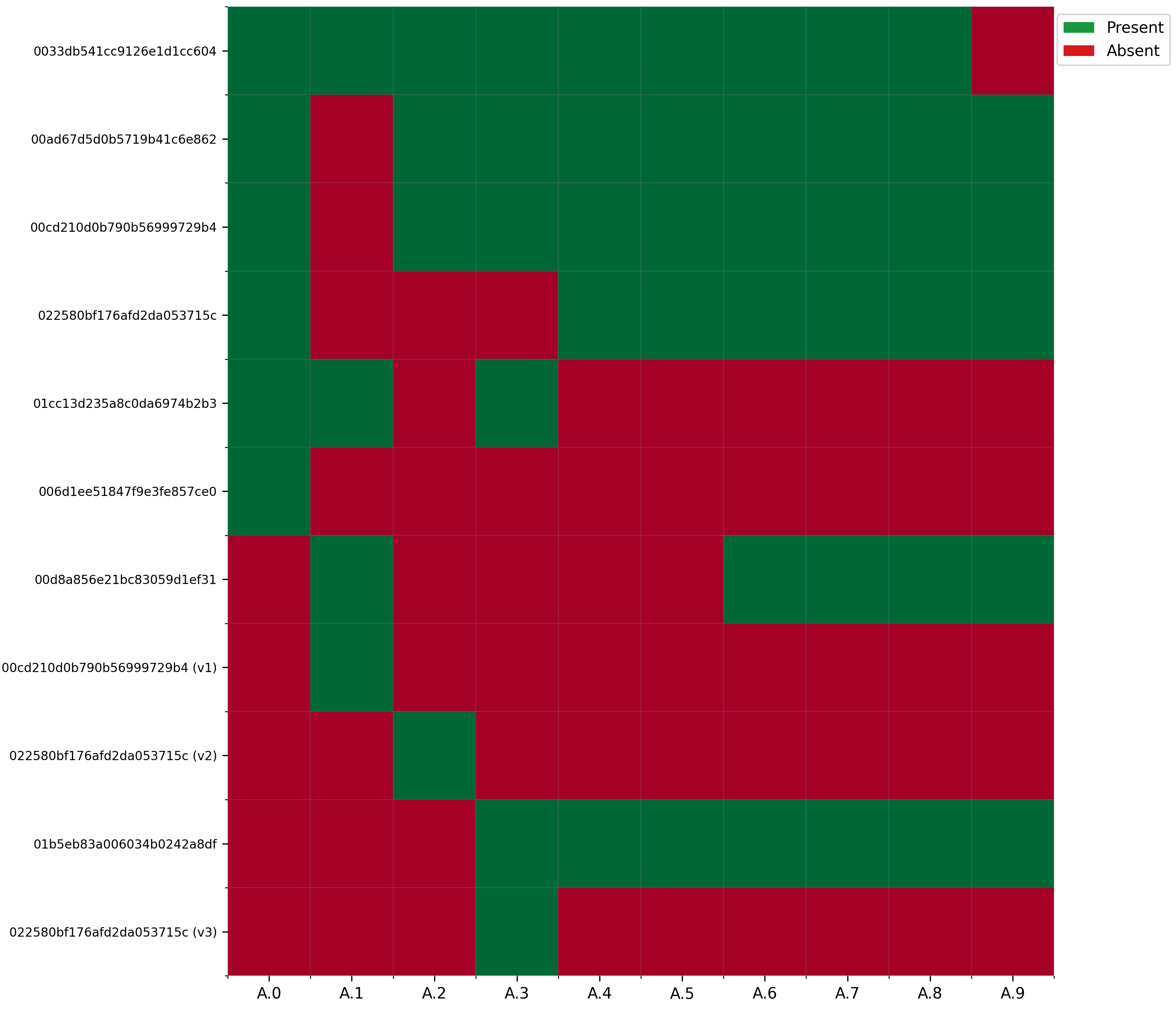}  
    \caption{Document presence matrix across SSL iterations. Rows correspond to pseudo-labeled documents (with version labels for variants), and columns represent SSL iterations. Green indicates inclusion, and red indicates exclusion.}  
    \label{fig:presence_matrix}  
\end{figure}

To illustrate the evolution of pseudo-labels during semi-supervised learning (SSL), we visualize all predictions on the unlabeled documents that were deemed sufficiently qualitative by the quality model. Out of approximately 20,000 predictions, only 4–6 documents per iteration were selected, resulting in a total of 11 unique pseudo-labeled documents across all iterations. Figure~\ref{fig:presence_matrix} shows a matrix plot where rows represent these documents (including their variants), and columns represent SSL iterations (A.0 through A.9). Green indicates inclusion, while red indicates exclusion.

\begin{figure}[h]  
\centering  
\begin{subfigure}[t]{0.32\textwidth}  
\centering  
\includegraphics[width=\textwidth]{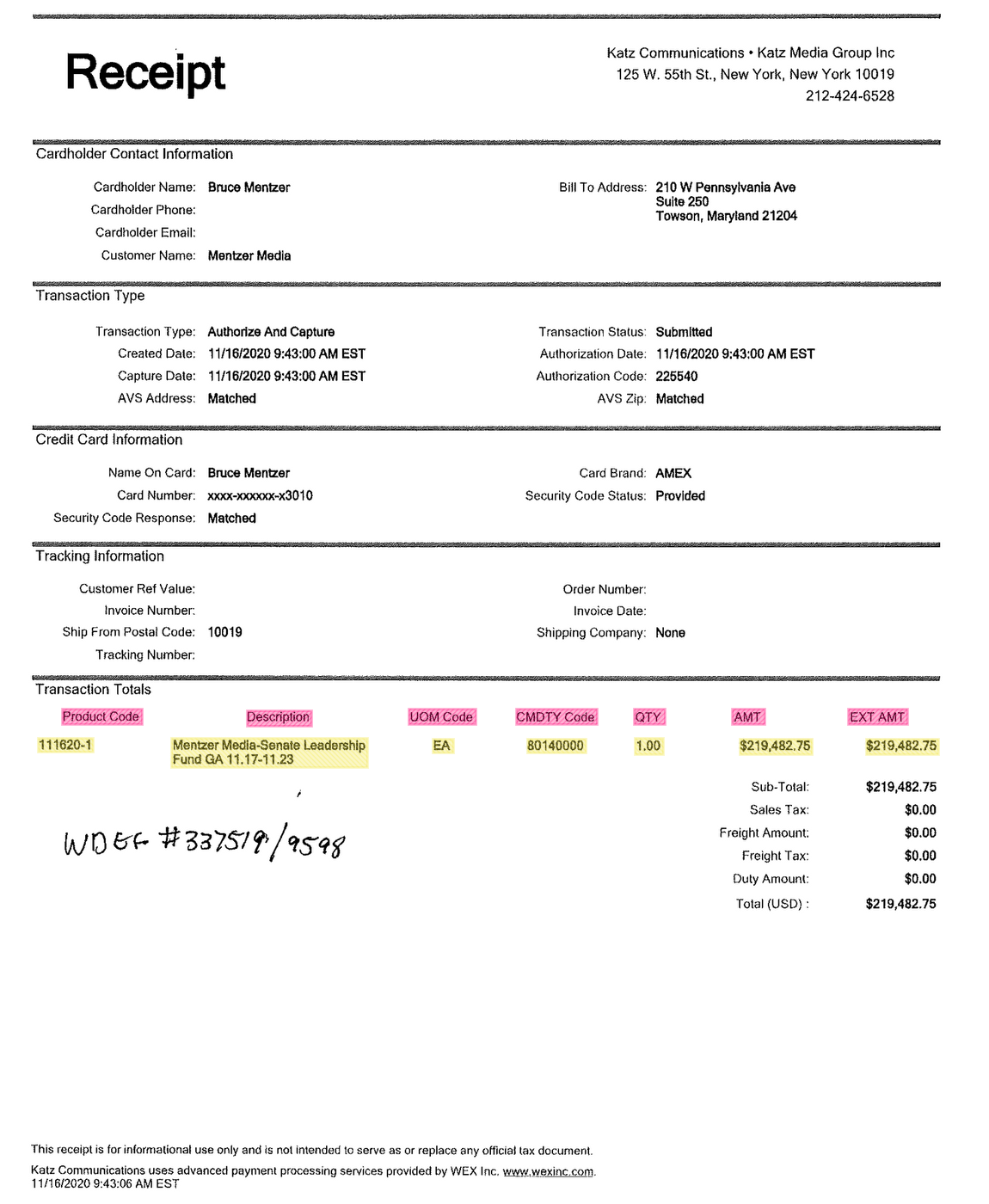}  
\caption{00ad6}  
\label{fig:doc1}  
\end{subfigure}  
\begin{subfigure}[t]{0.32\textwidth}  
\centering  
\includegraphics[width=\textwidth]{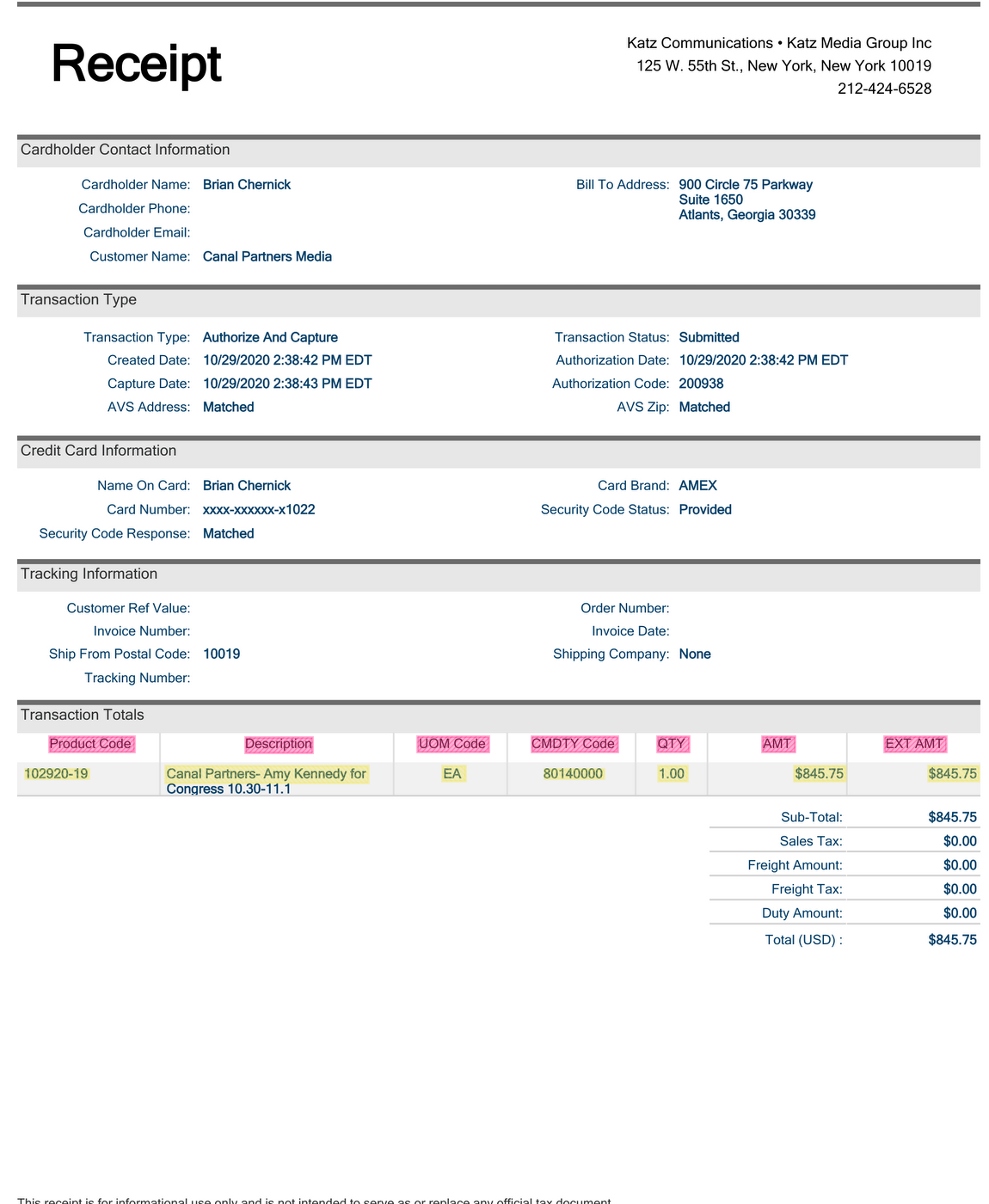}  
\caption{00cd2 (v0)}  
\label{fig:doc2}  
\end{subfigure}  
\begin{subfigure}[t]{0.32\textwidth}  
\centering  
\includegraphics[width=\textwidth]{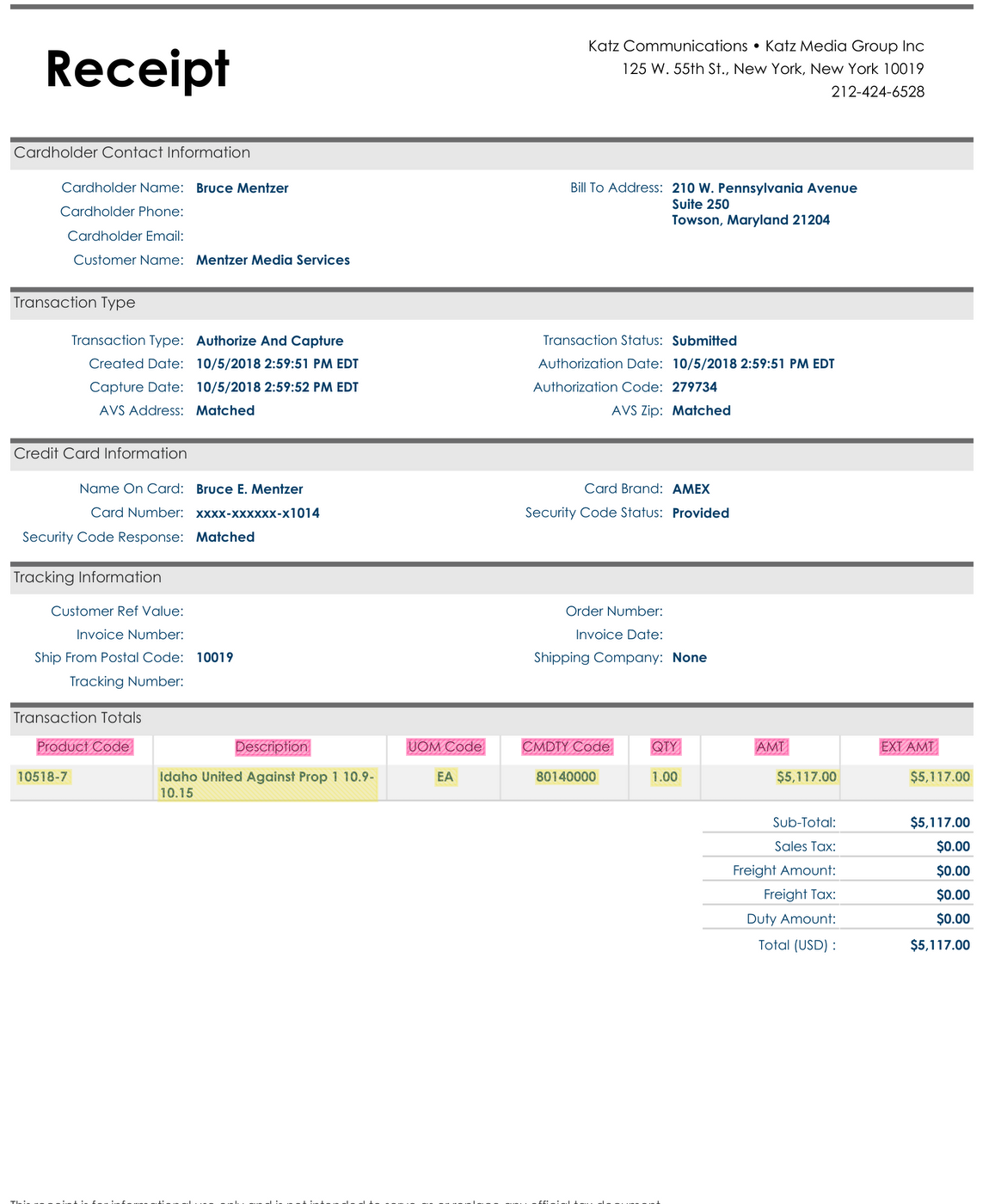}  
\caption{02258 (v0)}  
\label{fig:doc3}  
\end{subfigure}  
\caption{Documents that were kept throughout all iterations (it0 to it9). These represent consistently high-quality pseudo-labels.}  
\label{fig:group1}  
\end{figure}

\begin{figure}[h]  
\centering  
\begin{subfigure}[t]{0.32\textwidth}  
\centering  
\includegraphics[width=\textwidth]{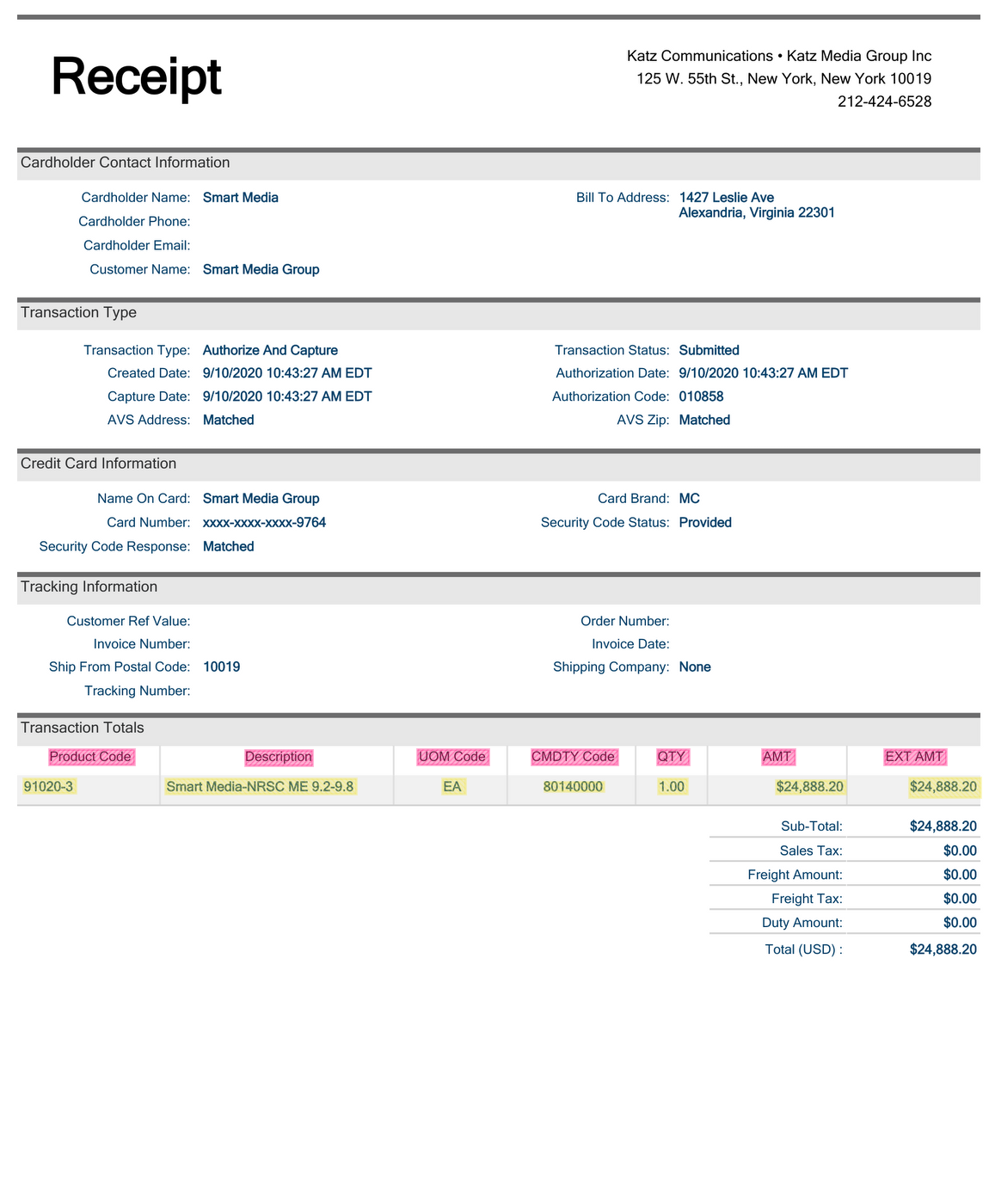}  
\caption{0033d}  
\label{fig:doc4}  
\end{subfigure}  
\begin{subfigure}[t]{0.32\textwidth}  
\centering  
\includegraphics[width=\textwidth]{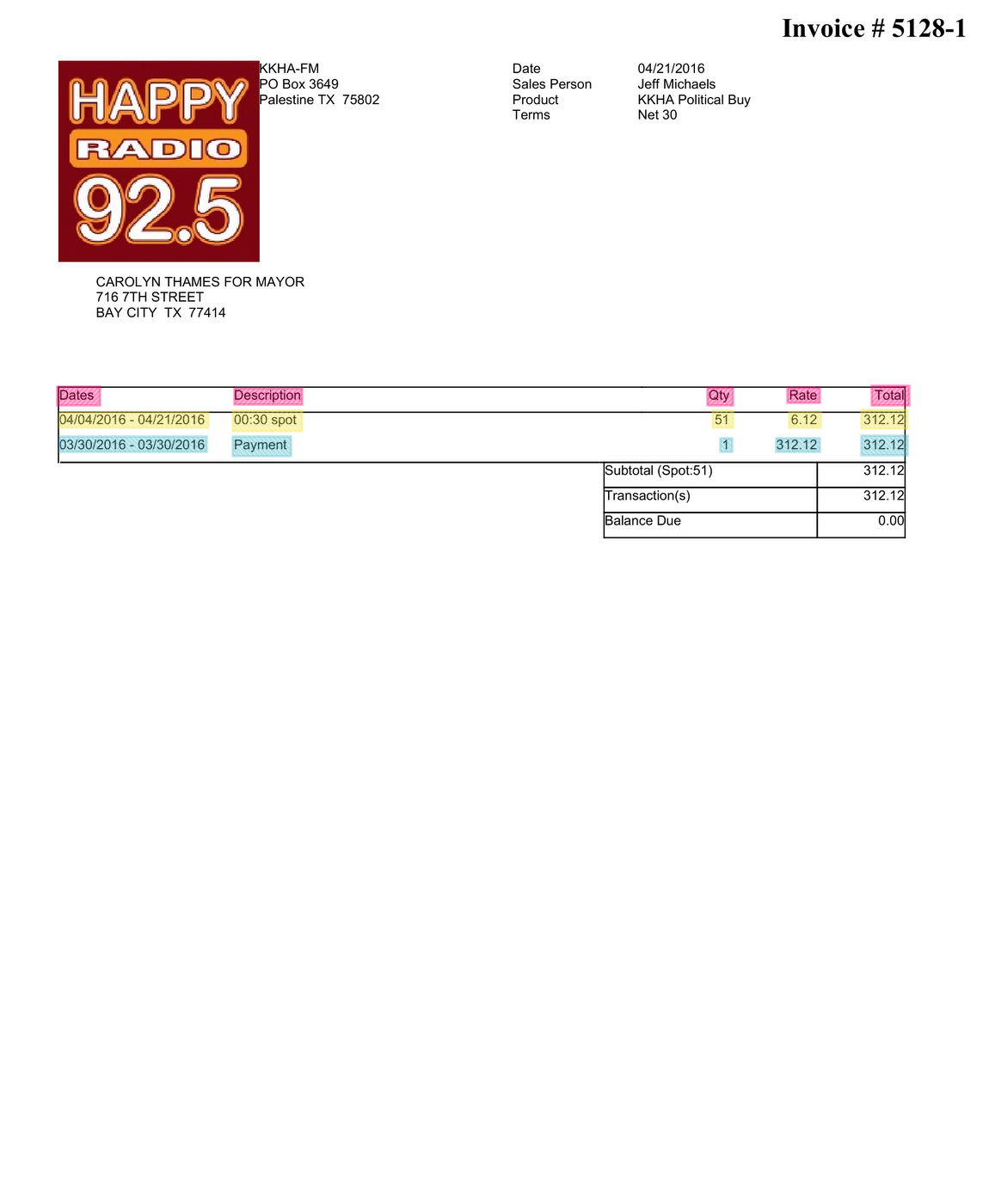}  
\caption{01cc1}  
\label{fig:doc5}  
\end{subfigure}  
\begin{subfigure}[t]{0.32\textwidth}  
\centering  
\includegraphics[width=\textwidth]{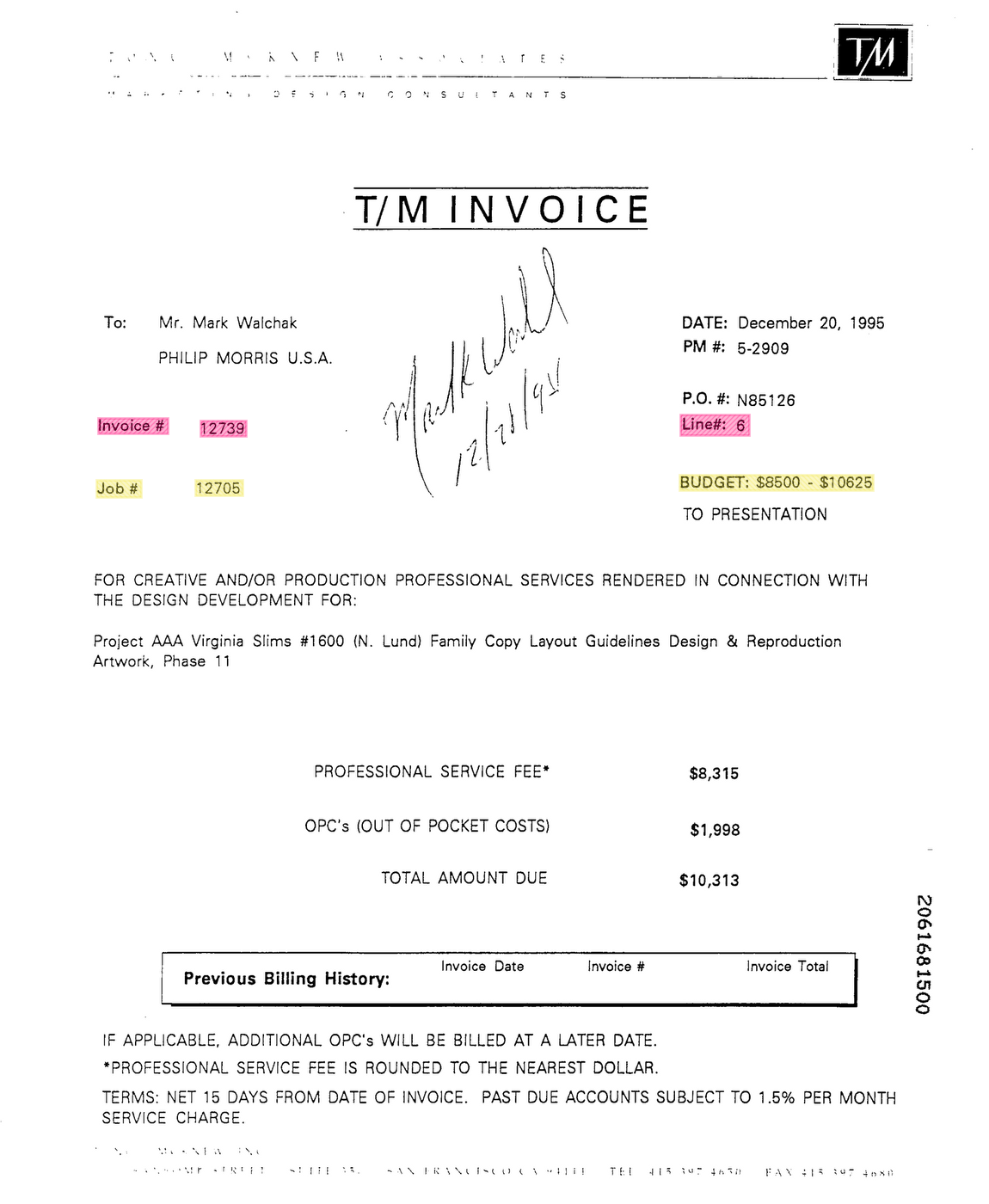}  
\caption{006d1}  
\label{fig:doc6}  
\end{subfigure}  
\caption{Documents initially included in it0 but removed in later iterations. While the pseudo-labels and predicted quality remained unchanged, the diversity check excluded these documents as they were too similar to others, offering limited additional information.  
 }  
\label{fig:group2}  
\end{figure}

\begin{figure}[h]  
\centering  
\begin{subfigure}[t]{0.32\textwidth}  
\centering  
\includegraphics[width=\textwidth]{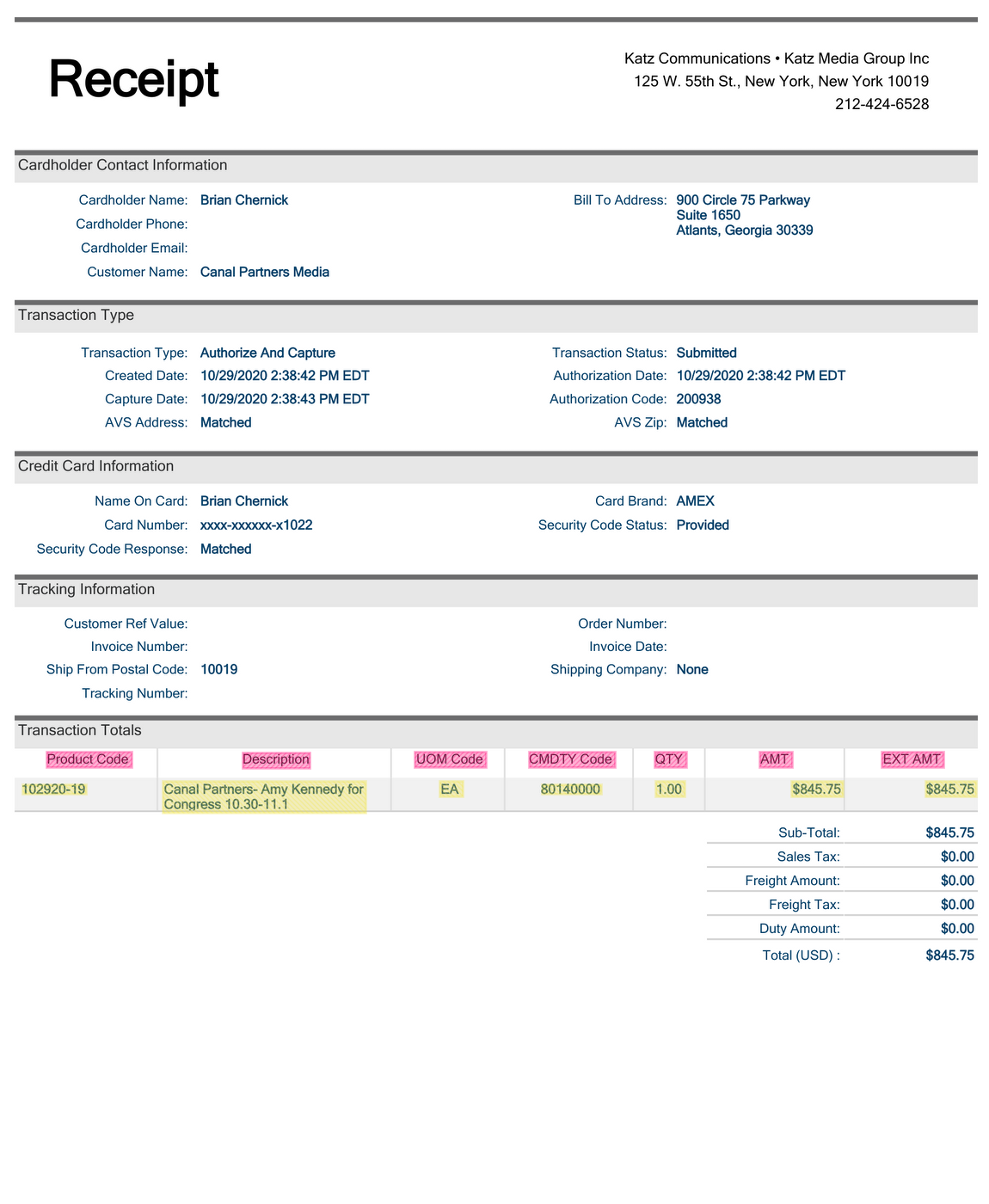}  
\caption{00cd2 (v1)}  
\label{fig:doc7}  
\end{subfigure}  
\begin{subfigure}[t]{0.32\textwidth}  
\centering  
\includegraphics[width=\textwidth]{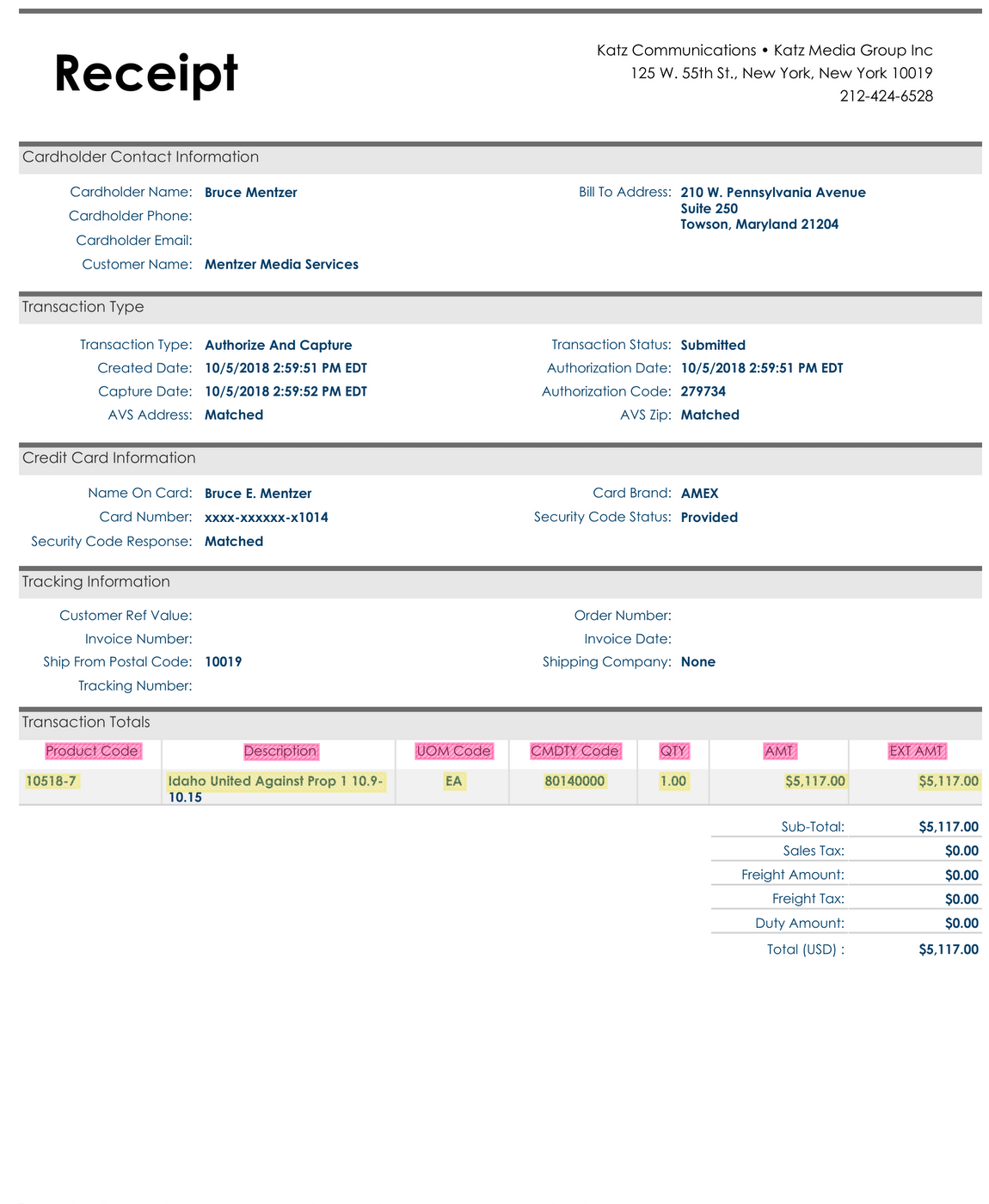}  
\caption{02258 (v2)}  
\label{fig:doc8}  
\end{subfigure}  
\begin{subfigure}[t]{0.32\textwidth}  
\centering  
\includegraphics[width=\textwidth]{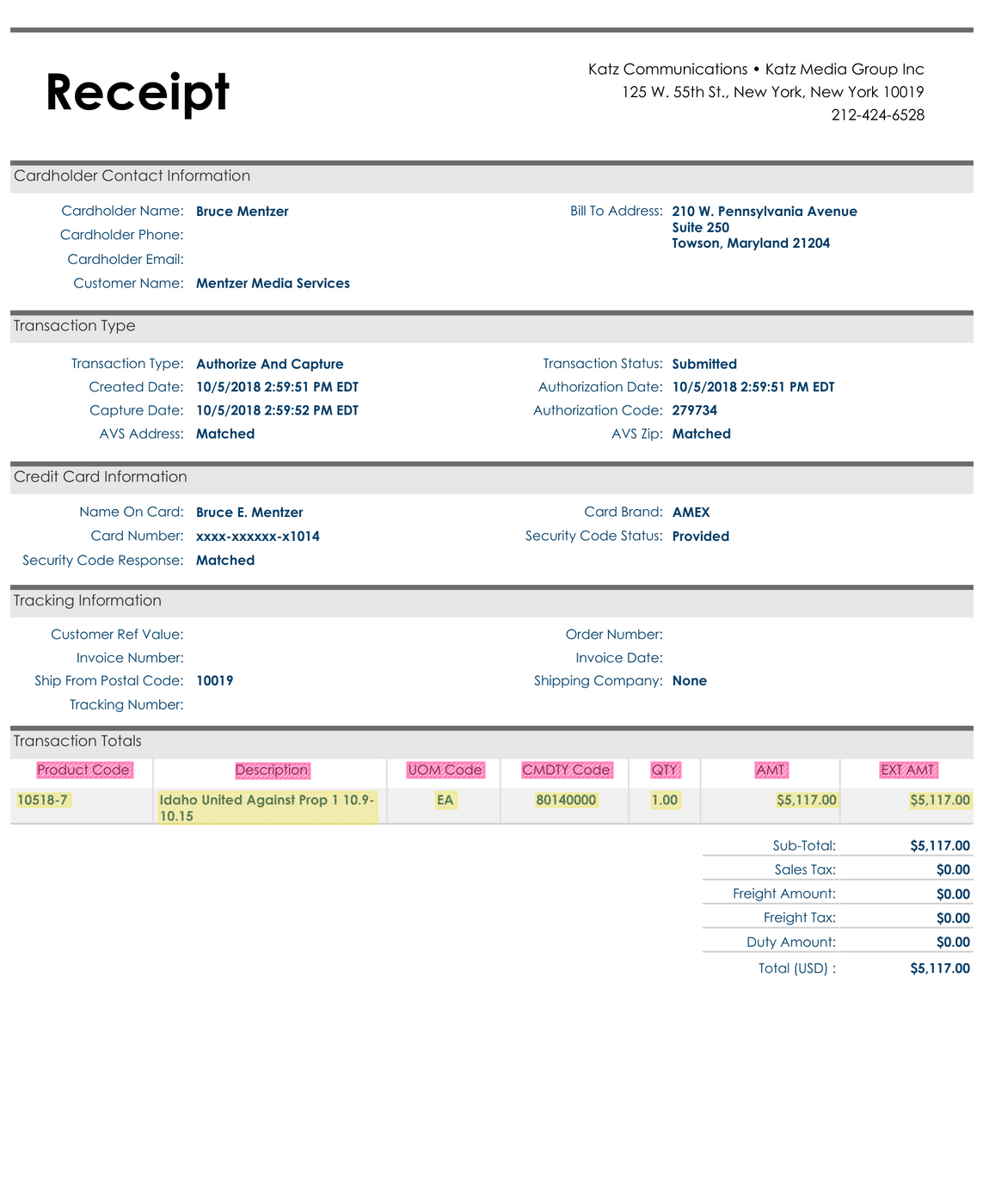}  
\caption{02258 (v3)}  
\label{fig:doc9}  
\end{subfigure}  
\caption{Documents added in later iterations (e.g., it1) but removed in subsequent iterations due to the diversity check, as they were too similar to others and offered limited additional information.  
}  
\label{fig:group3}  
\end{figure}

\begin{figure}[h]  
\centering  
\begin{subfigure}[t]{0.45\textwidth}  
\centering  
\includegraphics[width=\textwidth]{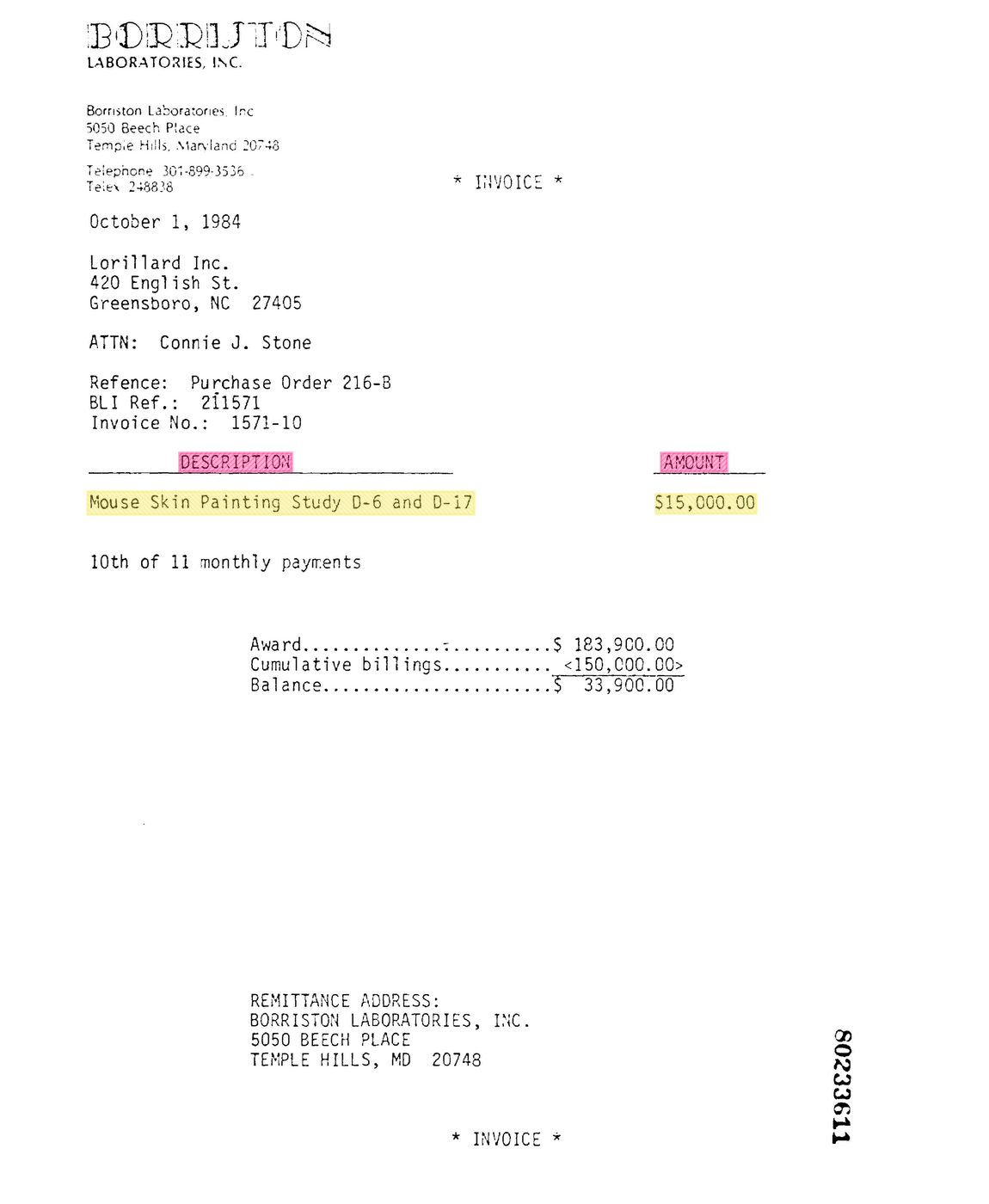}  
\caption{00d8a}  
\label{fig:doc10}  
\end{subfigure}  
\begin{subfigure}[t]{0.45\textwidth}  
\centering  
\includegraphics[width=\textwidth]{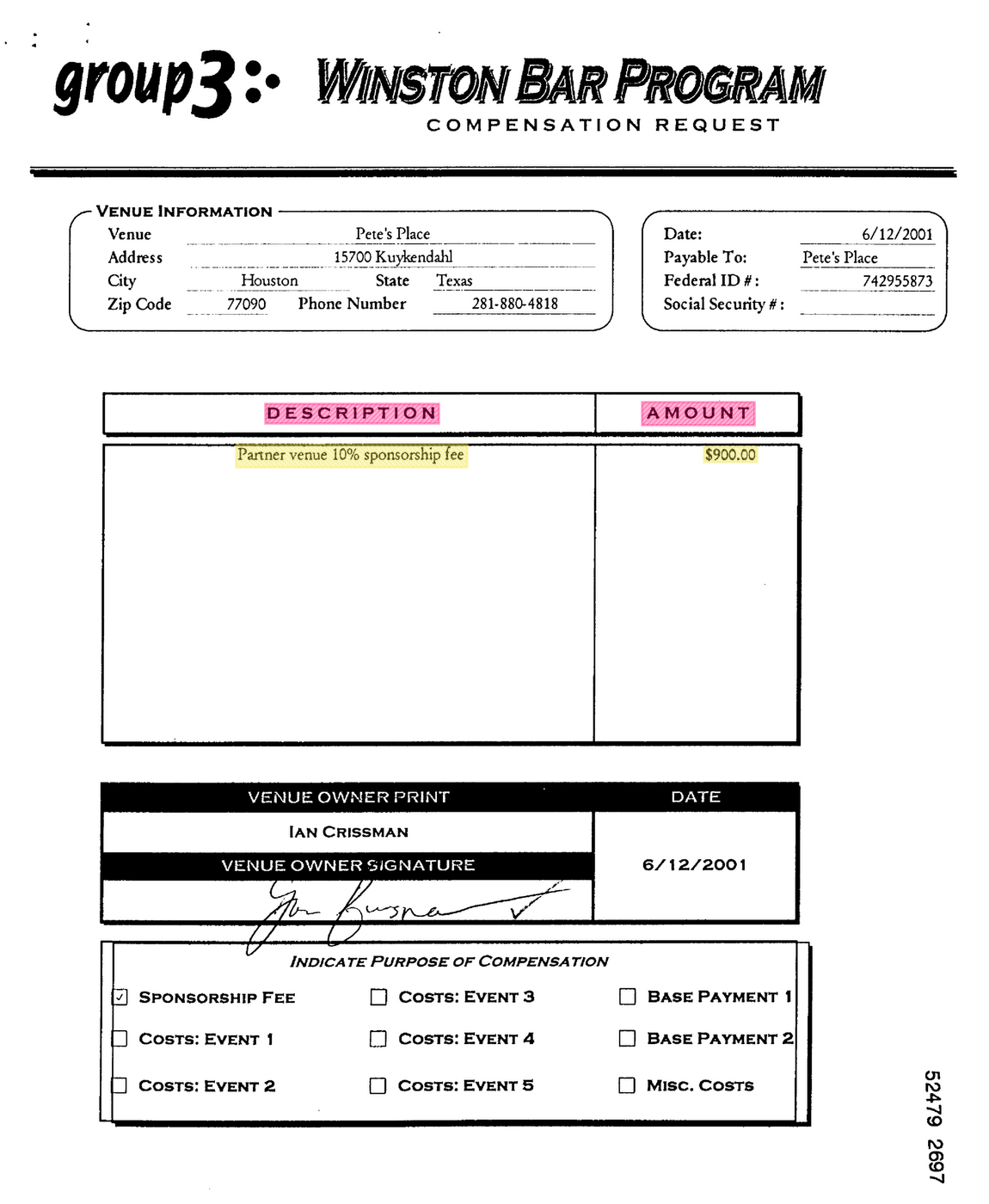}  
\caption{01b5e}  
\label{fig:doc11}  
\end{subfigure}  
\caption{Documents added in later iterations (e.g., it1) and retained until the final iteration. These represent cases where pseudo-labels became available due to the previous SSL iteration, and were ultimately used to train iteration 9, which produced the best extraction results.   }  
\label{fig:group4}  
\end{figure}

Figure~\ref{fig:group1} shows documents that were kept throughout all iterations (it0 to it9), representing consistently high-quality pseudo-labels. Figure~\ref{fig:group2} highlights documents that were initially included in it0 but removed in later iterations due to redundancy, as they were too similar to others and offered limited additional information.
Figure~\ref{fig:group3} illustrates documents that were added in later iterations (e.g., it1) but removed in subsequent iterations due to the diversity check. Finally, Figure~\ref{fig:group4} shows documents that were added in later iterations and retained until the final iteration, representing cases where pseudo-labels became available due to the previous SSL iteration and were ultimately used to train iteration 9, which produced the best extraction results. 

% Group 1: Documents kept from iteration 0 to iteration 9  

% Group 2: Documents from it0 that were not kept until it9  

% Group 3: Documents added later but not kept until it9  

% Group 4: Documents added later and kept until it9  

\end{document}